%% file: topk_paper.tex
\documentclass{article}

\usepackage{microtype}
\usepackage{graphicx}
\usepackage{booktabs} 

\usepackage[hyphens,spaces,obeyspaces]{url}
\usepackage{hyperref}

\usepackage[accepted]{icml2025_arxiv}

\usepackage{amsmath}
\usepackage{amssymb}
\usepackage{mathtools}
\usepackage{amsthm}
\usepackage{float}

\usepackage[capitalize,noabbrev]{cleveref}

\theoremstyle{plain}

\theoremstyle{definition}

\theoremstyle{remark}

\usepackage[textsize=tiny]{todonotes}
\usepackage{authblk}
\usepackage{subcaption, graphicx}
\usepackage{booktabs} 
\usepackage{xspace}
\usepackage{algorithm, algorithmic}

\usepackage[capitalize,noabbrev]{cleveref}
\usepackage[breakable]{tcolorbox}
\usepackage[inline]{enumitem}

\newtcolorbox[use counter=figure, 
crefname={figure}{figures}]%
{chatfig}[1][]{fonttitle=\bfseries,
title=Figure \thetcbcounter: #1}

\definecolor{cornflowerblue}{rgb}{0.39, 0.58, 0.93}
\definecolor{darkgreen}{rgb}{0.0, 0.5, 0.0}
\definecolor{lightblue}{rgb}{0.68, 0.85, 0.90}

\newcommand{\topk}{top-\texttt{k}\xspace}
\newcommand{\Topk}{Top-\texttt{k}\xspace}
\newcommand{\kk}{\texttt{k}\xspace}

\newcommand{\llama}{Llama}

\hypersetup{
    colorlinks=true,
    linkcolor=cornflowerblue,
    filecolor=magenta,      
    urlcolor=teal,
    citecolor=cornflowerblue,
    pdftitle={Long Context with Top-K Attention},
    pdfpagemode=FullScreen,
    }

\icmltitlerunning{Exploiting Sparsity for Long Context Inference}

\begin{document}

\twocolumn[
\icmltitle{Exploiting Sparsity for Long Context Inference:\\ Million Token Contexts on Commodity GPUs}



\icmlsetsymbol{equal}{*}

\begin{icmlauthorlist}
\icmlauthor{Ryan Synk}{equal,umd}
\icmlauthor{Monte Hoover}{equal,umd}
\\
\icmlauthor{John Kirchenbauer}{umd}
\icmlauthor{Neel Jain}{umd}
\icmlauthor{Alex Stein}{umd}
\icmlauthor{Manli Shu}{sales}
\icmlauthor{Josue Melendez Sanchez}{umd}
\\
\icmlauthor{Ramani Duraiswami}{umd}
\icmlauthor{Tom Goldstein}{umd}
\end{icmlauthorlist}

\icmlaffiliation{umd}{University of Maryland, College Park}
\icmlaffiliation{sales}{Salesforce Research}

\icmlcorrespondingauthor{Monte Hoover}{mhoover4@umd.edu}
\icmlcorrespondingauthor{Ryan Synk}{ryansynk@umd.edu}

\icmlkeywords{Machine Learning, Long Context}

\vskip 0.3in
]



\printAffiliationsAndNotice{\icmlEqualContribution} 

\begin{abstract}
There is growing demand for performing inference with hundreds of thousands of input tokens on trained transformer models. 
Inference at this extreme scale demands significant computational resources, hindering the application of transformers at long contexts on commodity (i.e not data center scale) hardware.
To address the inference time costs associated with running self-attention based transformer language models on long contexts and enable their adoption on widely available hardware, we propose a tunable mechanism that reduces the cost of the forward pass by attending to only the most relevant tokens at every generation step using a top-k selection mechanism. 
We showcase the efficiency gains afforded by our method by performing inference on context windows up to 1M tokens using approximately 16GB of GPU RAM. 
Our experiments reveal that models are capable of handling the sparsity induced by the reduced number of keys and values. 
By attending to less than 2\% of input tokens, we achieve over 95\% of model performance on common benchmarks (RULER, AlpacaEval, and Open LLM Leaderboard). The code is available at \href{https://github.com/ryansynk/topk-decoding}{https://github.com/ryansynk/topk-decoding}.
\end{abstract}

\input{Sections/introduction}
\input{Sections/motivation}
\input{Sections/methods}

\input{Sections/experiments}
\input{Sections/related_work}

\input{Sections/conclusion}


\bibliography{topk}
\bibliographystyle{icml2025}

\input{Sections/appendix}
\end{document}

%% file: Sections/introduction.tex
\section{Introduction}
\label{introduction}
\begin{figure}[!ht]
    \centering
    \includegraphics[width=0.9\linewidth]{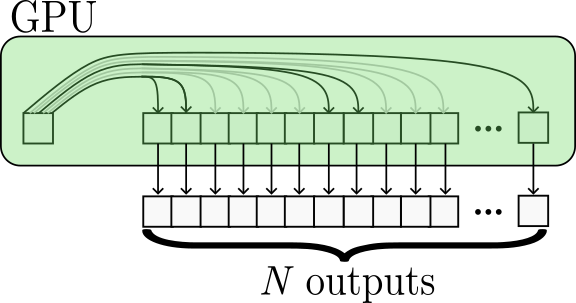}
    \includegraphics[width=0.9\linewidth]{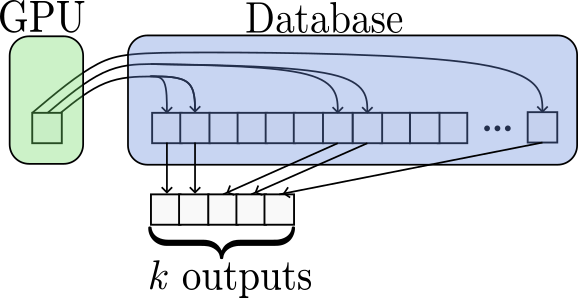}
    \caption{(top) Typical attention requires each query vector to compute an inner product with each key vector in the context window.  In practice, most key vectors produce insignificant attention scores, and therefore contribute very little to subsequent hidden states, so much of this computation is wasted. (left-bottom)  \topk attention retrieves only the keys that contribute significantly to the attention computation, leaving the gray arrows out and achieving sublinear runtime. }
    \label{fig:diagram}
\end{figure}

Long context inference is the process by which models analyze large document collections or follow long and detailed instructions. 
The \emph{context} is a series of tokens, like a set of documents or a set of files in a codebase. 
Increasingly, models are trained to handle larger and larger context lengths, with new models training on contexts in the millions \citep{liu2024worldmodelmillionlengthvideo}.

Inference is done using a trained model and is divided into two steps: \emph{prefill} and \emph{decoding}. In the prefill stage, tokens pass forward through the model, and their key and value activations from the attention mechanism of each layer are stored in a list. 
This list is called the \emph{KV cache} \citep{pope2023efficiently}. 
In the decoding stage, the model uses the KV cache to generate new tokens one at a time. 
In a standard attention mechanism, each new token attends to every cached token in the context. 
When the context length is very large, both the prefill and decoding stages can become prohibitively expensive.

The prefill stage incurs an $O(N^2)$ compute and memory cost due to the large attention matrix formed in the computation (where $N$ is the number of tokens in the context). To mitigate this, brute-force methods such as Ring Attention~\citep{liu2023ringattentionblockwisetransformers} were developed. This approach uses round-robin communication between servers to scale computation linearly in the number of GPUs. Despite its apparent effectiveness, while the prefill stage occurs only once, because the decoding stage occurs many hundreds of times for a single user request, the algorithm limited in its practicality outside of the data-center.

Every step during the decoding stage incurs an $O(N)$ compute and memory cost, as each new token must attend over all previous tokens in the cached context. As context lengths grow, the KV cache can become prohibitively large. If each of the $N$ tokens in the context is embedded into a vector of size $D$ (the \emph{hidden dimension}) then the number of floating point numbers in the cache is $2NDL$, where $L$ is the number of network layers. For a value of $D=4096$ and $L=32$ as in the Llama-3 8B architecture \citep{grattafiori2024llama3herdmodels}, at $N=100,000$ the memory required for the cache alone is 52GB, assuming 16 bit floating point format \citep{zhang2024pqcacheproductquantizationbasedkvcache}. At this context length the cache is unable to fit on most commodity GPUs. 

One approach to save memory during decoding is to offload the KV cache to CPU memory \citep{sheng2023flexgen}. However, the additional data movement required to implement this strategy is itself prohibitive. In the above example, just a single layer's worth of KV cache data is 1.6GB and this payload must be schlepped back and forth hundreds or thousands of times during a single generation. Attacking the problem from a different angle, cache eviction methods try to maintain a fixed cache size on the GPU by selectively removing token vectors deemed irrelevant \citep{zhang2023h2oheavyhitteroracleefficient}, but because it can be difficult to tell which vectors will be needed in the future, these strategies ultimately hurt performance.

\begin{figure}[t!]
    \centering
    \includegraphics[width=\linewidth]{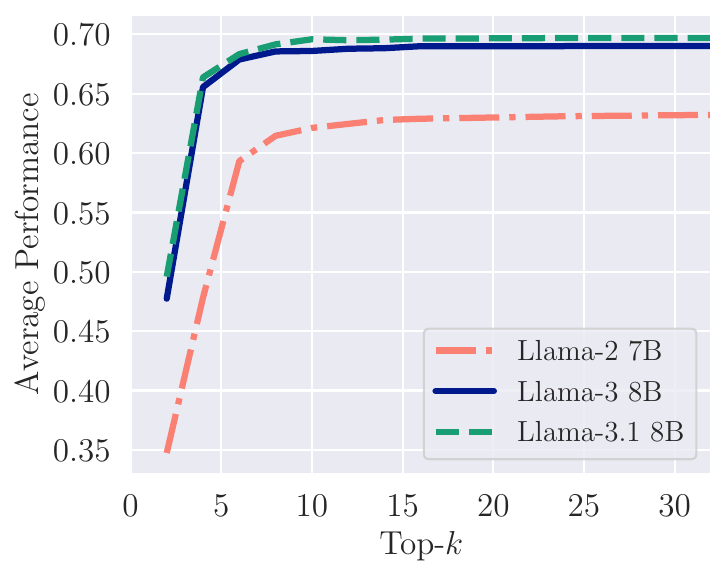}
    \caption{Performance on selected OpenLLM Leaderboard tasks using only the top-\texttt{k} keys for each attention computation. Typical questions have a context length of $\sim1000,$ yet only 10 keys are needed to achieve the same performance as full attention. We evaluate an extended set of models in \Cref{fig:lm_eval_harness_topk_results}.}
    \label{fig:lm_eval_harness_topk_results_teaser}
\end{figure}

In contrast to the compute and communication intensive solutions discussed thus far, we base our proposal on a simple fact: 

\begin{minipage}{1.0\linewidth}
\centering
\textit{Modern LLMs only need to pay attention\\to a handful of tokens at a time.}
\end{minipage}

We exploit this observation to perform fast long context decoding with very little GPU memory. 
We build an implementation of attention in which keys and values are stored in a vector database in CPU memory. 
When attention is computed using a query vector at decoding time, we retrieve only the \kk keys with the highest attention scores. 
This can be done in sublinear time using an approximate \kk-nearest neighbor search over the cache database, enabling long context inference using plentiful and cheap CPU memory, and without the heavy computational overhead required for full attention. Due to the fact that the constraint on the number of relevant keys constrains the CPU-GPU transfer volume, our approach incurs minimal data movement costs compared to other KV cache offloading methods since only a fraction of the token representations ever need to be transported between GPU and CPU memory.

To build a deeper understanding of why our approach works, through an array of controlled experiments we interrogate the core ``\topk'' assumption on which we base our method; that attention in modern LLMs is an inherently sparse operation. 
While methods exploring the restriction of the attention mechanism to the \topk scores have been studied extensively \citep{guptaMemoryefficientTransformersTopk2021, liu2024retrievalattentionacceleratinglongcontextllm, zhang2024pqcacheproductquantizationbasedkvcache}, 
prior work does not specifically endeavor to solve resource constrained long-context decoding using scalable tools like approximate nearest neighbor search, nor do they demonstrate million token context inference on single, commodity GPUs.

Our primary contributions are summarized as follows:
\begin{itemize}[topsep=0.0cm,itemsep=-0.2cm,leftmargin=0.5cm]
    \item We propose a simple method for sublinear long context generation using \topk attention over preprocessed states in a vector database.
    \item We show that this technique achieves high fidelity on common benchmarks and long-context tasks (\Cref{fig:lm_eval_harness_topk_results_teaser}).
    \item We provide an empirical analysis on why \topk attention works and demonstrate its effectiveness at the million token scale.
    \item We demonstrate the flexibility of our method by varying the \topk budget on a per-layer basis.
\end{itemize}

%% file: Sections/motivation.tex
\begin{figure}[t!]
    \begin{subfigure}[b]{0.48\columnwidth}
        \caption*{\hspace{6mm} Layer 1 of 32}
        \includegraphics[width=\linewidth]{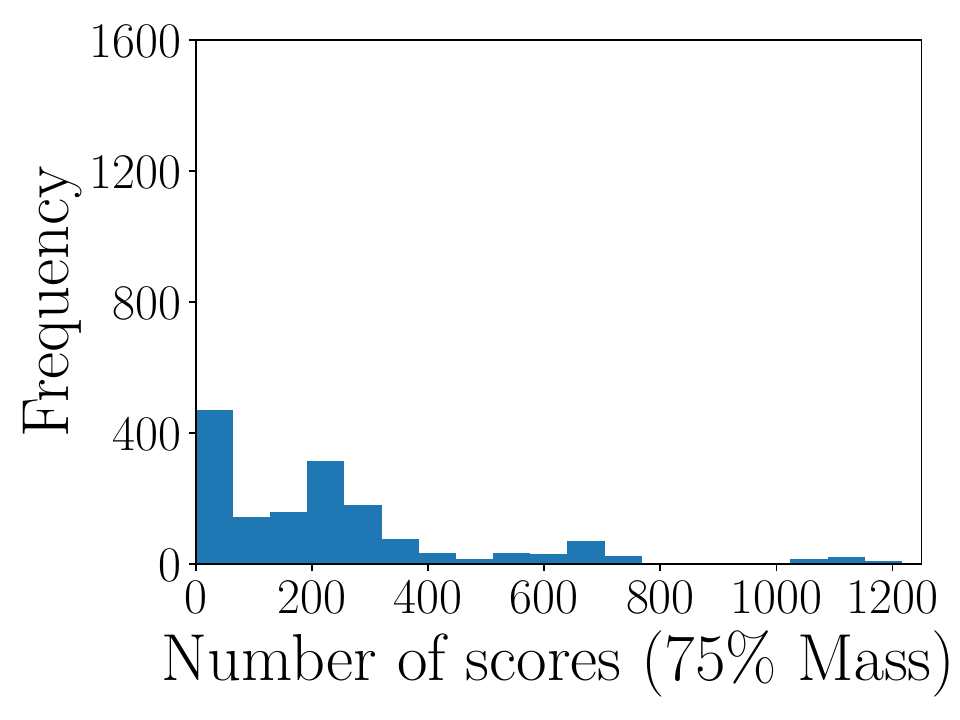}
    \end{subfigure}
    \hfill
    \begin{subfigure}[b]{0.48\columnwidth}
        \caption*{\hspace{6mm} Layer 16 of 32}
        \includegraphics[width=\linewidth]{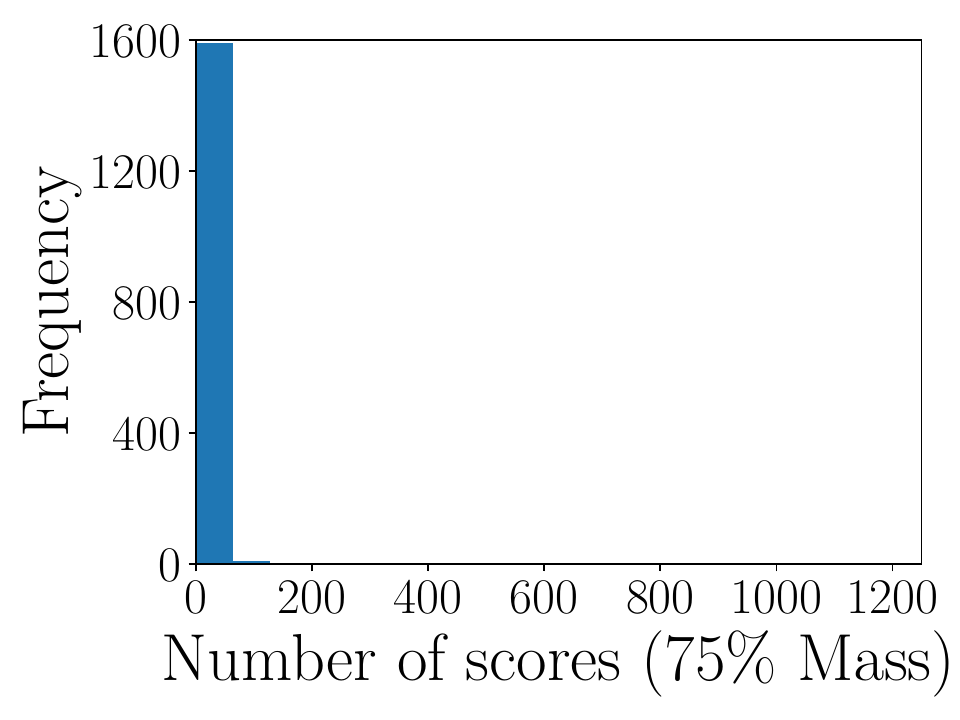}
    \end{subfigure}
    \caption{
    We analyze the number of attention scores that correspond to $75\%$ of the probability mass for generating the next token. Each point is the number of scores of the last `row' from the attention matrix required to reach $75\%$ of the total attention. We observe each of the $32$ heads across $50$ samples. 
    On the \textbf{left}, we plot the histogram for the first layer in the network, and on the \textbf{right} we plot it for layer 16. 
    }
    \label{fig:75_mass_hist}
\end{figure}

\section{Motivation}

The core assumption underpinning our proposal is the fact that modern language models naturally exhibit sparse attention patterns in which a very small number of tokens make up the vast majority of attention mass. We begin by verifying that this is in fact true for popular models in simple settings.

\paragraph{Observing attention sparsity in the wild.}

In \Cref{fig:75_mass_hist}, we analyze 50 4000-token text samples consisting of concatenated Wikipedia article snippets. We encode these samples using a forward pass through Llama-3-8B, and visualize the resulting attention scores.
For the last token in each context window, we tabulate the number of keys in the context that are needed to collect 75\% of the attention mass. 
Only a small number of the 4000 tokens are needed to collect this mass, especially for deeper layers of the networks.

\begin{figure}[t!]
    \centering
    \includegraphics[width=\linewidth]{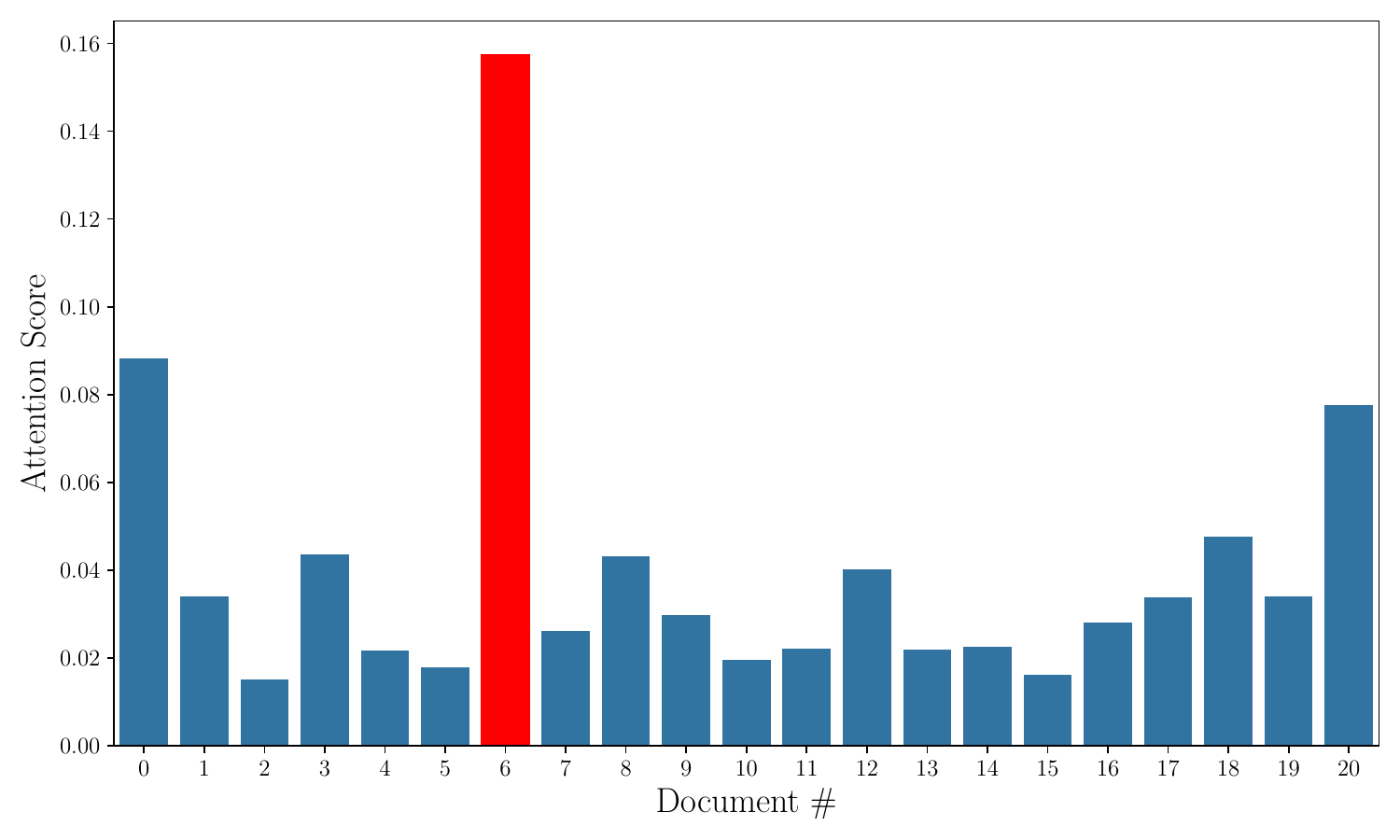}
    \caption{Average attention score per token in a given Wikipedia article for a multi-article long context sample (across all heads and layers). The bar for the article prompted to be copied is highlighted in red. Note how the model focuses its attention on the correct document.}
    \label{fig:attentionDoc}
\end{figure}

Next, we take these multi-article samples and prompt the model to copy one of the articles that relates to a specific topic. \Cref{fig:attentionDoc} demonstrates that in expectation the attention scores (across all heads and layers) are higher for the tokens contained within the correct document. 

\begin{figure}[h]
    \centering
    \includegraphics[width=\linewidth, trim= 2.0cm .2cm 2.5cm 1.8cm, clip]{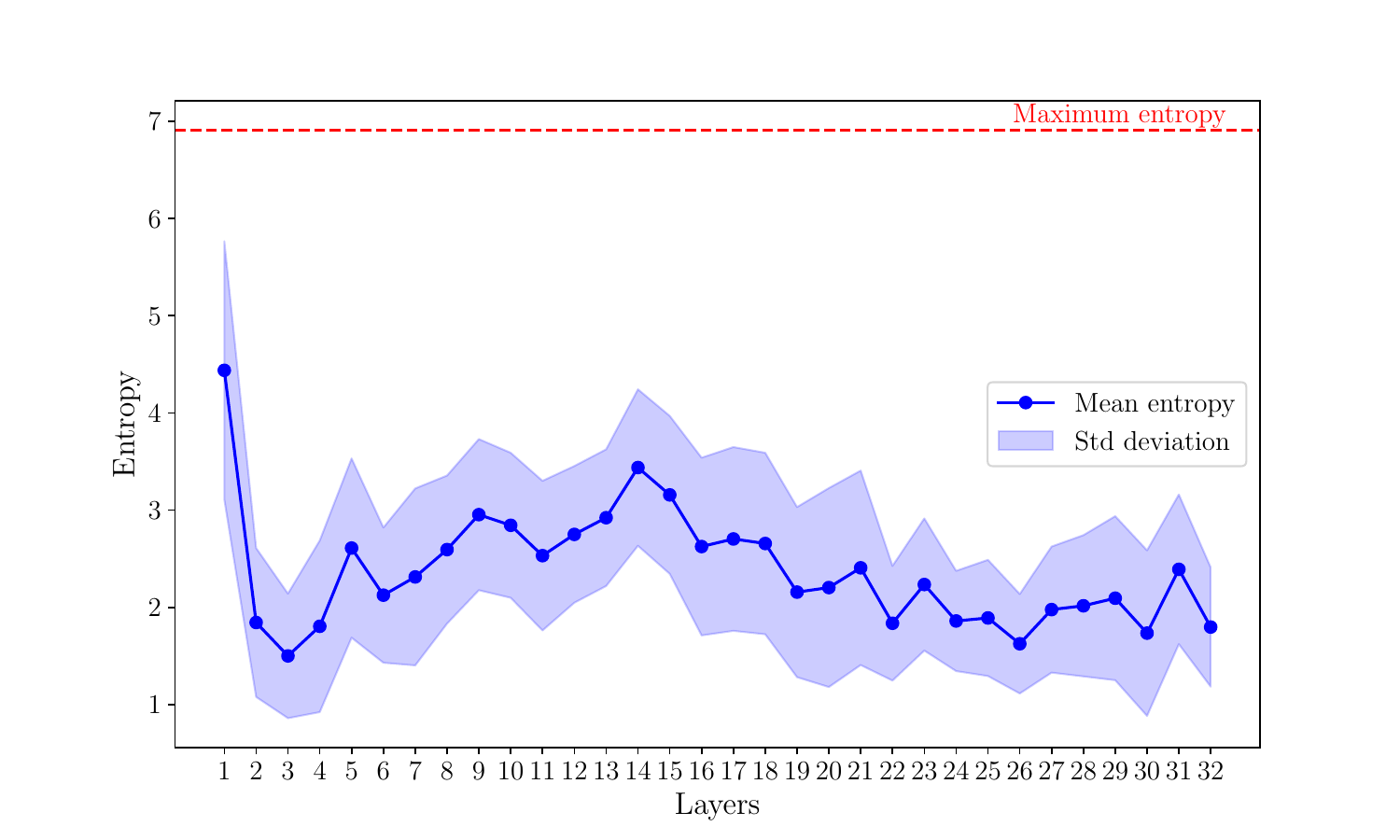}
    \caption{
    Entropy of the distribution of attention scores for each layer of a model, calculated as $E = - \sum_{i=1}^N a_i\log(a_i)$, where $(a_1,\dots,a_N)$ is the attention score distribution.  Attention score distributions are derived from last token of 50 1000-token samples and aggregated over all heads for a given layer. Entropy serves as a measure of how concentrated the attention scores are for a given query token: low entropy indicates a large amount of attention centered over few tokens, and high entropy indicates a more uniform dispersion of attention scores. Maximum entropy occurs when the distribution is perfectly uniform, and for 1000-token contexts is $-\log(\frac{1}{1000})$. 
    }
    \label{fig:entropy}
\end{figure}

Having confirmed that attention scores are indeed concentrated in terms of both how many tokens are relevant, and which tokens are highlighted in an input sequence, we zoom out and probe for structure in the attention scores in aggregate by quantifying their overall entropy. 
In \Cref{fig:entropy} we visualize the entropy of the last-token attention scores in a sequence as a function of layer depth. We observe that the entropy is low in all layers and decreases significantly after the first layer. Further analysis of attention sparsity across task categories and intuition on the connection between sparsity and entropy is provided in \Cref{app:attention_distribution}.

\paragraph{Is it safe to exploit attention sparsity?}

The observations made in this series of small scale experiments together suggest that
only a few key-value pairs should be needed for the model to perform close to its full capabilities. 
To demonstrate that the observed sparsity can actually be leveraged without compromising performance, we perform a more realistic test. 

We evaluate models from the Llama family on small selection of knowledge-intensive benchmark tasks while constraining the attention computation for each query to only use the \topk key-value pairs with the highest attention score (see \Cref{sec:topk-attention} for algorithmic details). In these experiments, we use the same number of keys for each layer of the network and average the model performance over all tasks on the OpenLLM leaderboard (see \Cref{sec:lm_eval_details} for task description). In \Cref{fig:lm_eval_harness_topk_results_teaser} we see that all three models saturate in performance by the 15th key, and the most recent (and most overtrained) variants of the model require only the 10 top keys to achieve the same performance as full-scale attentions. While \Cref{fig:75_mass_hist} indicates that there is some spreading of the attention mass for layer 1 of the network, this tail mass seems to be unnecessary for good performance on benchmarks.

Taken as a whole, these simple but motivating experiments evidence the basic sparsity assumptions on which our method is based. Empirically, we find that the attention scores are a native, reliable indicator as to which key-value pairs are most critical during any given inference step, and that very few keys are actually needed to perform accurate inference. 

In the next section we introduce the technical details of our approach for using a vector database to retrieve the \topk most influential tokens, thereby disconnecting attention's selection operations from the rest of the feed-forward computation. This critical separation enables the GPU to perform smaller matrix operations that fit into its limited memory whilst the full cached keys and values live in CPU memory and are searched using CPU compute. In the main experiments that follow, we showcase how our method alleviates both the computational and memory barriers that normally bottleneck long-context inference and unlocks million token contexts for small GPUs.

%% file: Sections/methods.tex
\section{Methodology}
\label{methods}

In the following sections we elaborate on our method for reducing memory costs during inference time using \topk attention. Along the way, we more concretely define some minimal notation and terminology required to describe inference operations in transformer language models.

\subsection{Inference With KV Caches}

As stated previously, causal inference using trained transformer models is split into two steps: the prefill and decode stages. Assume we are given a---potentially very large---set of tokens $x$ of size $N$ that we would like to make many queries over.

The query, key and value embeddings during prefill are of size $(N\times D)$, so this stage of inference incurs a maximum activation memory cost of $O(N^2)$ due to the size of the attention matrix. Memory saving attention methods like \cite{dao2022flashattentionfastmemoryefficientexact} can in practice reduce this to an $O(N)$ memory cost by tiling the attention computation. As the cache is being filled, it also grows in size and persistently costs $O(N)$ memory across the duration of the computation.

In the decoding stage, we have a smaller query we would like to make over this context, as well as a pre-filled KV cache. Our queries comes in one token at a time, with decoding in a self-attention layer performed as follows:

\begin{equation}\label{eqn:attn}
    \text{Attention}(q,K,V) = \text{Softmax}\left(\frac{qK^{T}}{\sqrt{D}} \right)V
\end{equation}

Here $q \in \mathbb{R}^{1\times D}$ is the new query vector, $, K, V \in \mathbb{R}^{N\times D}$ are the KV cache for that layer, and $D$ is the hidden dimension. We refer to $S = qK^T \in \mathbb{R}^{1\times N}$ as the \emph{score} matrix, and the quantity $\text{Softmax}\left(\frac{S}{\sqrt{D}} \right)$ as the \emph{attention} matrix.

When the incoming query $q$ is multiplied by all previous keys in the cache during decoding, the memory and compute cost incurred for the score matrix at this step is $O(N)$. This dependence on $N$ for the cost of GPU memory during inference can become prohibitive for large context lengths.


\subsection{\Topk Attention: Accelerating Decoding}\label{sec:topk-attention}

We cut down on the growing memory and compute costs by dynamically considering only the most relevant keys in the KV cache. To do this we perform a \texttt{k}-nearest neighbor search over the key vectors with the new query vector, returning only those value vectors corresponding to the \texttt{k}-largest score values.

The core of our implementation is in \Cref{alg:gen_topk}. We assume we are given a prefilled KV cache that is on the CPU. This cache is a sequence $\{K_{\ell}, V_{\ell}\}_{\ell=1}^{L}$ of key and value activation tensors, one for each layer. 

First we convert the key tensors across all layers in the KV cache into a nearest-neighbor search index. This index data structure can be as simple as a list (for exact nearest neighbors) or as sophisticated as a graph-based structure (for approximate nearest neighbors) as in \cite{10.1109/TPAMI.2018.2889473}. For multi-head attention, we construct a separate index for each key head at every layer. For each layer we store the indices in a list called $\texttt{K\_index}$.

We embed each new token during decoding into $q, k, $ and $v$ on the GPU. We offload $q$ to the CPU and perform a nearest neighbor search to get the \topk score values of the query over the context:
\[
    \texttt{vals}, I = \texttt{knn\_search}\left(q, \texttt{K\_index}[\ell], \texttt{k}\right)
\]
Here \texttt{k} is the number of the largest score values we'd like use. The operation $\texttt{knn\_search}$ performs a \texttt{k}-nearest neighbor search of $q$ over the key vector index for layer $\ell$, $\texttt{K\_index}[\ell]$. Here the \texttt{k}-nearest neighbor search distance is measured using the dot product metric, mirroring the attention score mechanism. This search returns two items: \texttt{vals}, a vector of size $\texttt{k}$ containing the \topk score values, and $I$, a list of $\texttt{k}$ integers mapping the score values in $\texttt{vals}$ to their corresponding key vectors. Let $I = (i_1, \dots i_k)$. We collect the relevant keys selected by the nearest neighbor search by concatenating them as row vectors together, denoted as $V_{\ell}[I]$:
\[
    V_{\ell}[I] =
    \begin{pmatrix}
        \text{---} \hspace{-0.2cm} & V_{\ell}[i_1] & \hspace{-0.2cm} \text{---}\\
        \text{---} \hspace{-0.2cm} & V_{\ell}[i_2] & \hspace{-0.2cm} \text{---}\\
                                   & \vdots        &                           \\
        \text{---} \hspace{-0.2cm} & V_{\ell}[i_k] & \hspace{-0.2cm} \text{---}\\
    \end{pmatrix}
    \in \mathbb{R}^{k\times D}
\]

After the model generates the first token using the context, the key and value embeddings of this token are left on the GPU directly. This splits the total KV cache into two parts: a large part on the CPU constructed from the given context, and a small part on the GPU constructed from previously generated tokens. Given a new query, the \texttt{k}-nearest neighbor search is only performed on the CPU cache; on the GPU, we compute the attention directly between this query and the keys from previous generated tokens. This process resembles windowed attention \citep{Beltagy2020Longformer}, where a window of recently generated tokens are stored directly on the GPU. These GPU window caches are labeled $\texttt{K\_gen}$ and $\texttt{V\_gen}$ in \Cref{alg:gen_topk}.

The $V_{\ell}[I]$ and \texttt{vals} are then moved to the GPU. Using these, we perform the final attention computation, combining both the weighted values from the context as well as the value vectors from previously generated tokens. This method gives us a peak GPU memory cost of $O(k)$ GPU memory cost, as opposed to $O(N)$. We find that $k$ can be chosen to be a small fraction (around 1\% for most tasks) of $N$ while still recovering near-equivalent performance.

\begin{algorithm}
    \caption{\Topk KV Cache Decoding}\label{alg:gen_topk}
    \begin{algorithmic}[1]
        \REQUIRE KV-Cache $\{K_{\ell}, V_{\ell}\}_{\ell=1}^{L}$ where $K_{\ell}, V_{\ell} \in \mathbb{R}^{N \times D}$, token $x\in \mathbb{R}^{1\times D}$, $\texttt{k} \in \mathbb{N}$
        \STATE $N_{gen} = 1$, $\texttt{K\_gen} = []$, $\texttt{V\_gen} = []$
        \STATE \texttt{K\_index} = []
        \FOR{$\ell \in \{1, \dots, L\}$}
            \STATE $\texttt{K\_index}[\ell] \gets \texttt{build\_index}(K_{\ell})$
        \ENDFOR 
        \WHILE{$N_{gen} < N_{max}$}
            \FOR{$\ell \in \{1, \dots, L\}$}
                \STATE Pre-attention transformer layer computations...
                \STATE $q = xW_{Q_{\ell}}$, $k = xW_{K_{\ell}}$, $v = xW_{V_{\ell}}$
                \STATE $\texttt{K\_gen}[\ell] \gets \texttt{concat}(\texttt{K\_gen}[\ell], k)$  
                \STATE $\texttt{V\_gen}[\ell] \gets \texttt{concat}(\texttt{V\_gen}[\ell], k)$ 
                \STATE $\texttt{vals}, I \gets \texttt{knn\_search}(q, \texttt{K\_index}[\ell], \texttt{k})$ 
                \STATE Move $V_{\ell}[I]$, \texttt{vals} to GPU
                \STATE $\hat{x}\gets \texttt{Softmax}(\frac{1}{\sqrt{D}}\texttt{vals})V_{\ell}[I] $ 
                \STATE $\hat{x}\gets \hat{x} + \texttt{Softmax}\left(\frac{1}{\sqrt{D}}q\cdot\texttt{K\_gen}[\ell]^T\right)$
                \STATE Post-attention transformer layer computations...
            \ENDFOR
            \STATE $x \gets \texttt{sample\_new\_token}(\hat{x})$
            \STATE $N_{gen} \gets N_{gen} + 1$
        \ENDWHILE
    \end{algorithmic}
\end{algorithm}
\subsection{Prefilling a KV Cache at the Million Token Scale}
The prefill forward pass in the construction of a KV cache potentially requires incurring the full $O(N^{2})$ memory cost. However, there are multiple ways one could prefill a cache. 

Given (relatively brief) access to large amounts of compute, one could parallelize over many GPUs and construct the cache using algorithms like Ring Attention \cite{liu2023ringattentionblockwisetransformers}. The larger up-front cost of the cache construction would then be amortized over the many queries that would be made over it on much cheaper hardware. Additionally, the attention could be approximated in the construction of the cache. Algorithms like windowed attention would allow for the construction of large caches with more modest compute as in \cite{child2019generatinglongsequencessparse}. For small enough $N$ (100's of thousands of tokens), and with a high-memory GPU, the vLLM library can quickly construct KV caches by performing a standard forward pass on the model using the paged attention algorithm \cite{kwon2023efficient}. Finally, for very large N, \topk attention could be employed at cache construction time as well. 

For our largest experiments, we utilize Flash Attention and an H100 GPU to prefill the caches $N$=1M tokens of context. This allows us to generate exact prefilled KV caches with only small modifications to the network to accommodate the extreme memory requirements. The exact changes we made in order to prefill caches can be found in the Appendix and we employ a chunking strategy similar to \cite{guptaMemoryefficientTransformersTopk2021}. We note that our experiments mimic an important use case of our method, namely a user with a large amount of documents that has access to only a limited amount of compute. In this case, the user can rent the required compute on the cloud for prefill \emph{once}, then make as many queries as they want quickly with our method on their own hardware.

%% file: Sections/experiments.tex
\section{Evaluating \Topk Attention at Scale}
\label{experiments}

We evaluate \topk to highlight the relationship between different values of \kk and performance. We find that models of various sizes and generations perform well even at small \kk. In general, we observe that a \kk equal to 2\% of the total length of the context is sufficient to achieve 95\% of the performance achieved with full, standard attention.

\subsection{Benchmark Details}
We analyze the effectiveness of \topk attention across three benchmarks: RULER, Open LLM Leaderboard v1, and AlpacaEval. Each of these benchmarks highlight a different aspect of \topk attention impact on LLM performance. RULER specifically tests long context abilities and we use Open LLM Leaderboard v1 and AlpacaEval together to examine how different model sizes and families perform under \topk attention. Measuring performance across these benchmarks presents a comprehensive understanding of how the \topk attention mechanism affects a model. 

\paragraph{RULER} To demonstrate that \topk attention with small \kk remains effective as the context length increases, we run the RULER \citep{hsiehRULERWhatsReal2024} benchmark suite over a series of increasing context lengths. As shown in \Cref{tab:ruler_raw_results} we run over lengths from $4$k to $128$k. 
RULER consists of thirteen total tasks from four categories. The evaluation harness runs the original Needle In A Haystack (NIAH) \citep{kamradt2023needle} task along with a series of more challenging variations of the task (for example, in one task the text consists entirely of labeled ``needles" and the model is queries to retrieve the needle corresponding to a single label.) These NIAH tasks comprise $8$ of the $13$ tasks, and the remaining tasks are split into three categories: summarization proxies, multi-hop proxies, and question answering. 

\paragraph{Open LLM Leaderboard Tasks}\label{sec:lm_eval_details} We investigate the performance of \topk on Open LLM Leaderboard tasks. In particular, we evaluate different models on various values of \kk on the average of MMLU \citep{hendrycks2020measuring-MMLU}, ARC tasks both easy and challenge \citep{clark2018arc}, HellaSwag \citep{zellers-etal-2019-hellaswag}, winogrande \citep{ai2:winogrande}, OpenbookQA \citep{OpenBookQA2018}, BoolQ \citep{clark2019boolq}, and PiQA \citep{Bisk2020}. For each task, we record the normalized accuracy when available; otherwise, we record accuracy. We report the average over tasks. We evaluate the following models on these benchmarks: \llama-1 (7B), \llama-2 (7B), \llama-3 (8B), \llama-3.1 (8B), Vicuna-v1.3 (7B), \llama-2 chat (7B), \llama-3 Instruct (8B), \llama-3.1 instruct (8B), \llama-3.2 1B instruct, and \llama-3.2 3B instruct \citep{touvron2023llama, zheng2023judging, touvron2023llama2, dubey2024llama, Meta2024llama3_2}. The experiments are conducted using \texttt{lm-eval-harness} in a zero-shot setting \citep{eval-harness}. 

\paragraph{AlpacaEval 2.0} We benchmark \topk attention on AlpacaEval \citep{dubois2024length}. AlpacaEval $2.0$ requires a model to generate responses to $805$ queries. These responses are then compared by an LLM-as-a-Judge with GPT-4 Turbo responses generated from the same query set. The winrate percentage is the reported metric, and the LLM-as-a-Judge is GPT-4 Turbo. 

\subsection{Evaluation Results}


\paragraph{\Topk Performance on RULER} 
We evaluate our method on RULER using GradientAI's Llama-3-8B model that was trained using a context length of 262k tokens. For this model, we find that very small values of \kk are sufficient to recover near-baseline performance. At every context length evaluated, $95\%$ of the baseline performance can always be achieved with a \kk value of $1\%$ or less of the total length. In \Cref{tab:ruler_raw_results}, at $k=2$, greater than $60\%$ performance is achieved at all context lengths. The performance on RULER improves as \kk increases. Nevertheless, even at a context length of $131$k tokens to achieve $\sim 98\%$ performance of the full attention only $12.5\%$ of the attention scores are required.

\begin{table}[b!]
\centering
\caption{Results on RULER benchmark at various context lengths. Scores represent an average of 13 tasks in the RULER benchmark, with maximum possible score being 100. The RULER benchmark was run separately for each context length listed, and each context length was run with \topk attention for increasing values of \kk and also with standard, full attention.}\label{tab:ruler_raw_results}
\begin{tabular}{l|ccccc}
\toprule
& \multicolumn{5}{c}{\textbf{Context Length (tokens)}} \\ 
\textit{k}              & {8192} & {16384} & {32768} & {65536} & {131072} \\
\midrule
\textbf{2}              & 70.58                & 71.27                 & 68.00                 & 67.83                 & 64.76                  \\
\textbf{8}              & 83.10                & 86.69                 & 84.92                 & 75.99                 & 61.28                  \\
\textbf{32}             & 85.38                & 88.08                 & 85.20                 & 78.18                 & 65.89                  \\
\textbf{128}            & 87.21                & 89.08                 & 85.16                 & 77.41                 & 73.59                  \\
\textbf{512}            & 88.55                & 89.31                 & 84.56                 & 77.41                 & 63.58                  \\
\textbf{2048}           & 89.42                & 89.31                 & 84.53                 & 77.33                 & 64.62                  \\
\textbf{8192}           & --                   & 88.85                 & 84.90                 & 77.41                 & 58.53                  \\
\textbf{16384}          & --                   & --                    & 84.81                 & 77.26                 & 73.40                  \\
\textbf{32768}          & --                   & --                    & --                    & 78.03                 & 73.40                  \\
\textbf{Full}           & 90.31                & 89.45                 & 85.03                 & 78.87                 & 75.17  \\ \bottomrule            
\end{tabular}
\end{table}

\begin{figure*}[ht]
    \centering
    \includegraphics[width=0.32\linewidth]
    {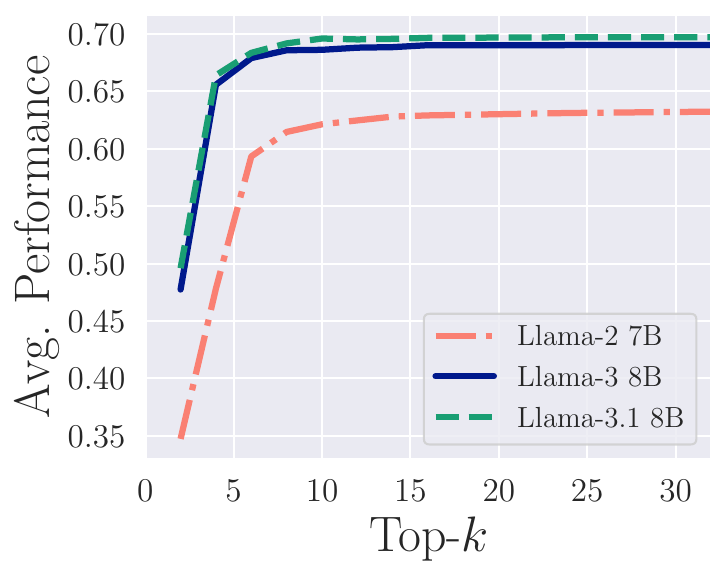}
    \includegraphics[width=0.32\linewidth]{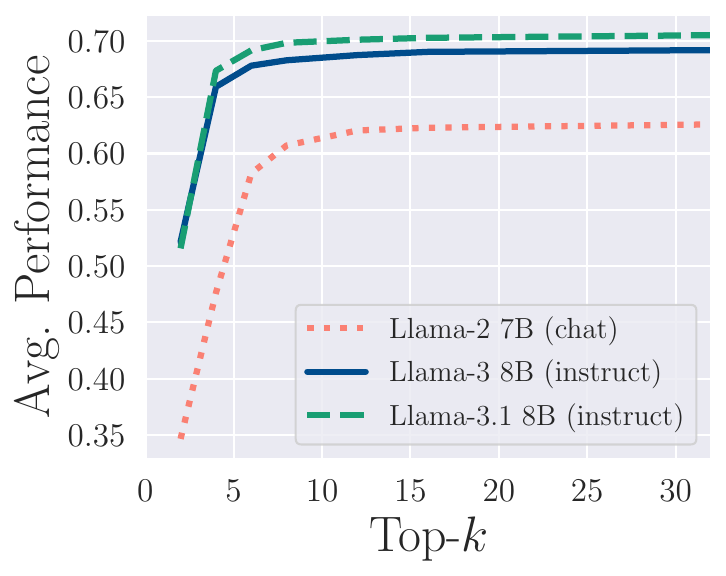}
    \includegraphics[width=0.32\linewidth]{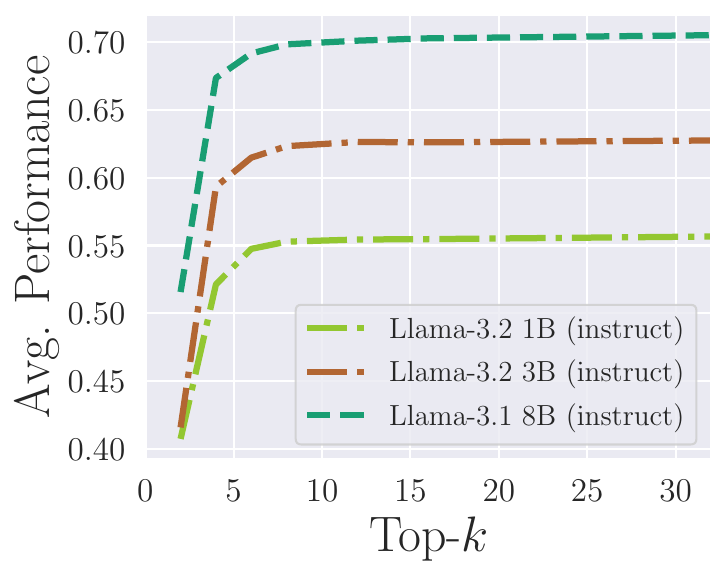}
    \caption{\Topk attention is effective for OpenLLM Leaderboard Tasks even at small values of \kk. Left shows the average of all tasks as we increase \kk on pretrained base models. Center shows instruction tuned models. Right investigates the performance on different model sizes.}
    \label{fig:lm_eval_harness_topk_results}
\end{figure*}

Interestingly, NIAH tasks have high success rates with little variation across choice of \kk. Question-answering tasks (from SQuAD \citep{rajpurkar2016squad100000questionsmachine} and HotpotQA \cite{yang2018hotpotqadatasetdiverseexplainable}) have lower success rates with the 8B-size model, but also show low variation in success rates across different \kk. In our experiments, word-counting tasks (CWE and FWE) were the tasks most affected by the choice of \kk. This suggests that in practice, extra compute (larger \kk) should be allocated to these types of tasks to maintain performance. \Cref{tab:entropy_correlation} shows the differences between each of these tasks in terms of what \kk-value is required to reach near-parity performance. 

\paragraph{\Topk Performance on OpenLLM Leaderboard Tasks} We evaluate various models to find that \topk behavior exists regardless of instruction tuning, model size, or the number of tokens the model was trained on. \Cref{fig:lm_eval_harness_topk_results} left shows that regardless of the number of tokens on which the model was trained, all exhibit a similar trend with performance saturating at \kk values $<10$. When comparing \Cref{fig:lm_eval_harness_topk_results} left and center, we see that instruction models and pretrained base models exhibit similar behavior, saturating very quickly. Finally, in \Cref{fig:lm_eval_harness_topk_results} right, while different models achieve different performance maximums on the benchmark suite, we see that the effect of \topk is independent of model size.

\begin{figure}[htbp]
    \centering
    \includegraphics[width=0.75\linewidth]{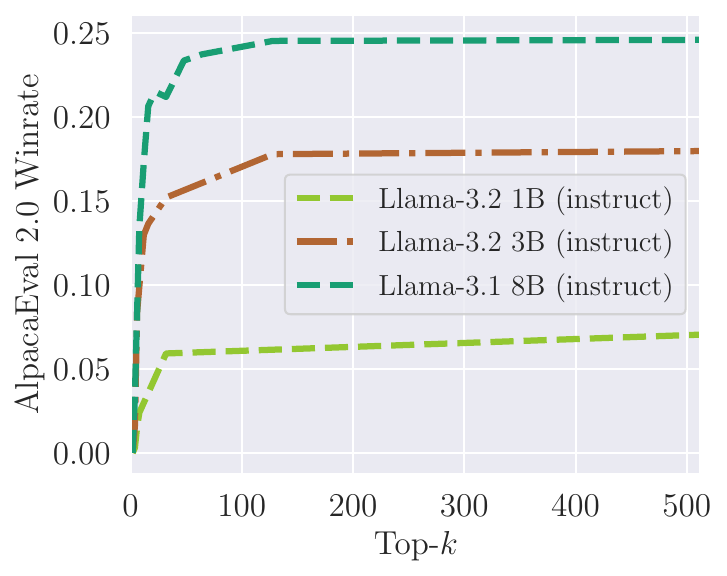}
    \caption{\kk equivalent to 2\% of context length is sufficient to achieve 95\% of dense-attention performance on AlpacaEval 2.0, regardless of model size.}
    \label{fig:alpaca_eval_topk_results}
\end{figure}

\paragraph{\Topk Performance on AlpacaEval}
\Cref{fig:alpaca_eval_topk_results} shows that the \textit{95\% performance with 2\% of attention} result holds on generation-intensive tasks as measured by AlpacaEval 2.0. The trend of small \kk values achieving near-baseline performance holds true across model sizes, and additional results showing performance across model generations can be found in the appendix (\Cref{fig:alpaca_eval_topk_results_app}).

\begin{figure}
    \centering
    StreamingLLM \\
    \includegraphics[width=0.9\linewidth]{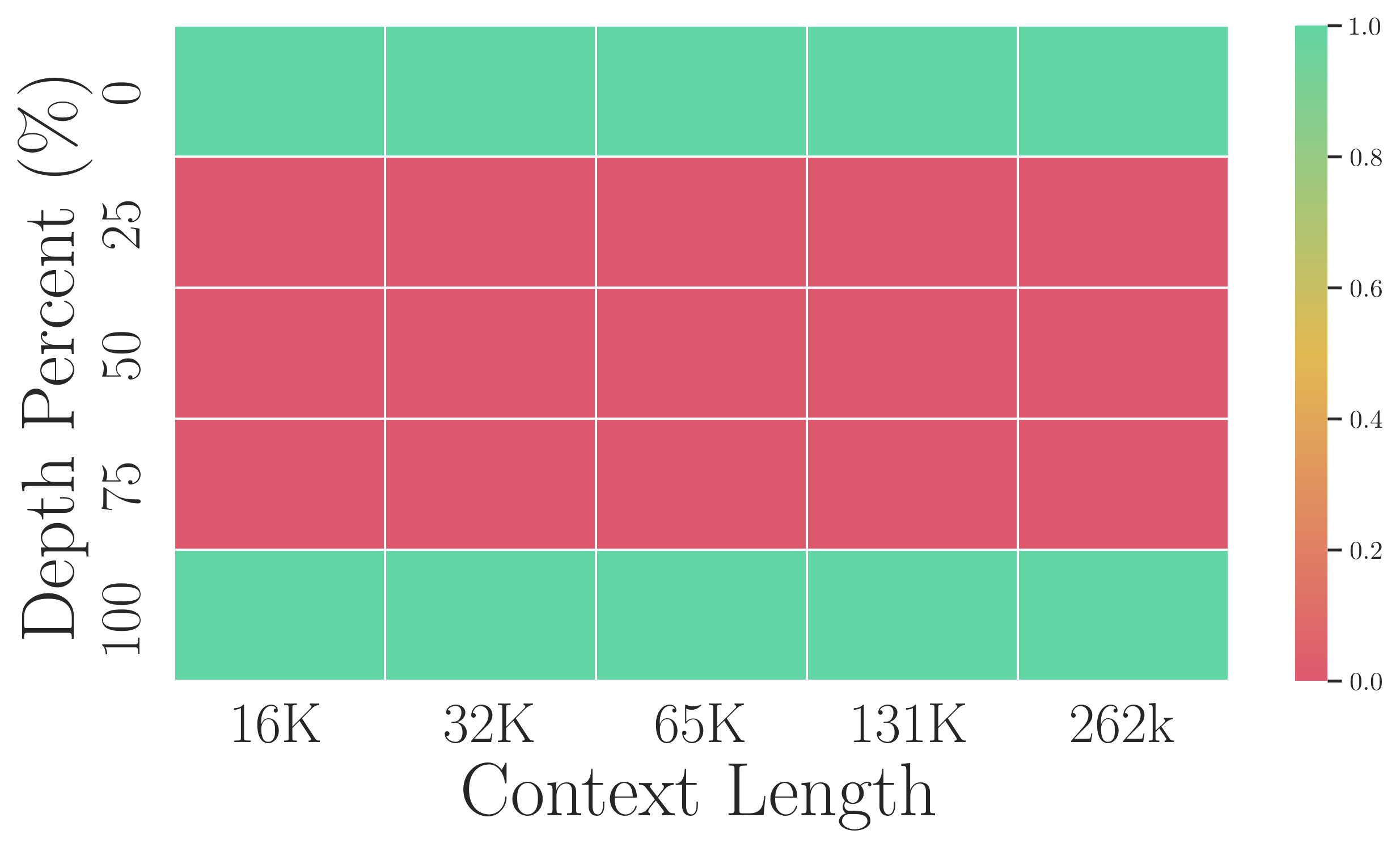} \\
    Ours \\
    \includegraphics[width=0.9\linewidth]{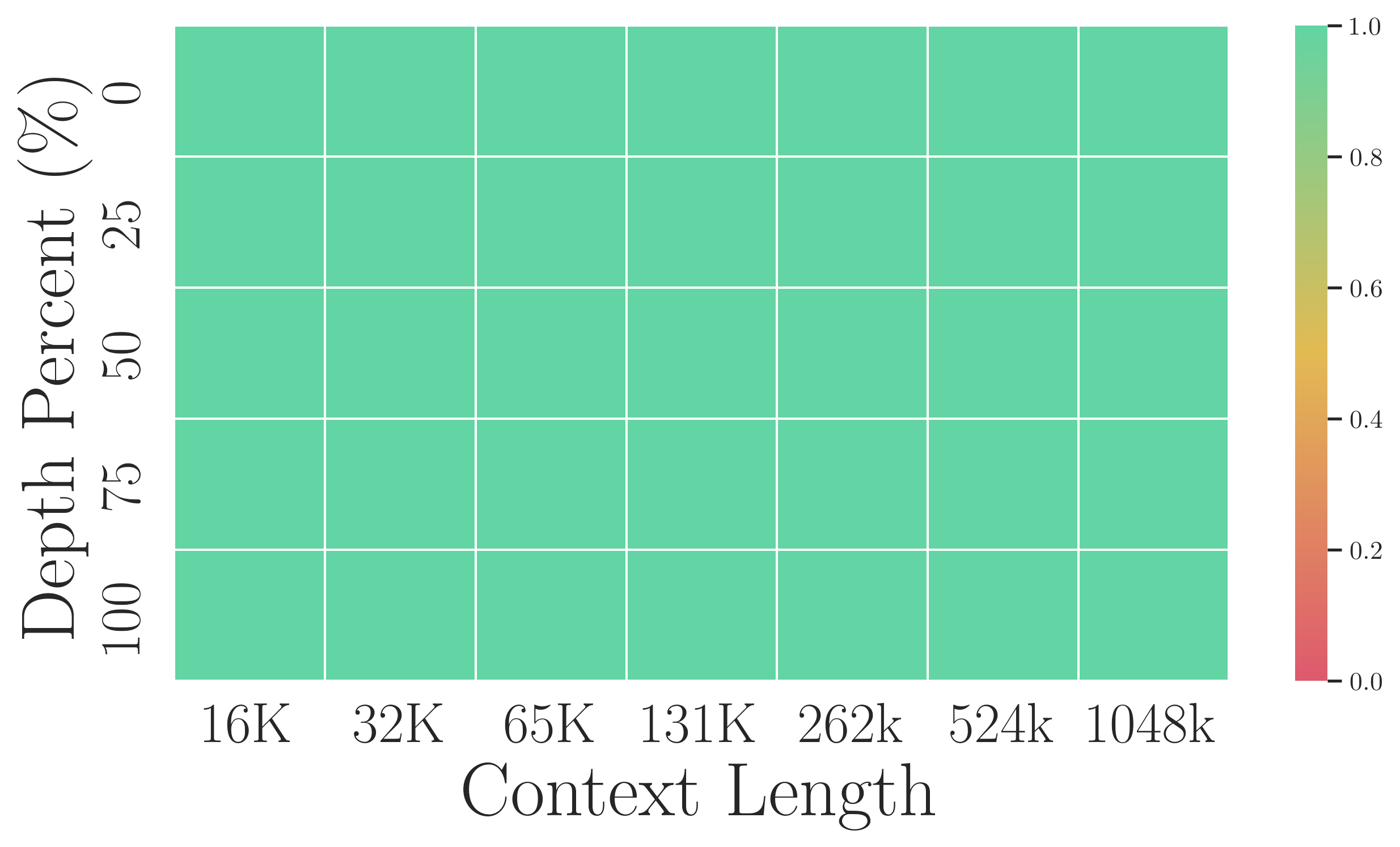}
    \caption{One million token NIAH performance comparing cache eviction \cite{xiao2023streamingllm} and \topk attention. The red cells show that attention sinks are incapable of retrieving tokens outside of the local window or early sink tokens. \Topk attention achieves 100\% success with just $k=10$.}
    \label{fig:niah_plots}
\end{figure}

\subsection{1M Token Generation with \Topk}
To demonstrate the extreme scaling that \topk attention permits, we use our method to generate tokens conditioned on a context window of \textit{one million} tokens. We choose RULER's Needle In A Haystack task to use for the context, and we use GradientAI's Llama-3-8B model that was trained out to a context length of 1M \citep{gradient2024scaling}. We run this experiment on a single GPU with a Faiss vector database prefilled with the contents of a KV cache.\footnote{We use a variety of \kk values for this experiment find that \kk=1 is sufficient to solve Needle In A Haystack with a context length of 1 million tokens.} We also compare our method against \cite{xiao2023streamingllm} for needle in a haystack in \Cref{fig:niah_plots}. Note that \topk attention has a 100\% success rate regardless of where in the context the needle is hidden, which cannot be said for any available cache-eviction method.

\subsection{Exploring Non-Uniform Choices of \kk}

In this final section we report the results of our studies on the minimal \kk required for strong performance across different task types and on the impact of applying our \topk operator non-uniformly over the layers of the transformer.

\paragraph{Optimal \kk Across Task Types}
Table \ref{tab:entropy_correlation} shows how the value of \kk necessary to achieve 95\% of base model performance varies across the tasks in RULER. For the needle in a haystack task, only a tiny fraction of the entire context is necessary to achieve 95\% of the base performance of the model, whereas nearly 9\% is necessary in the case of Word Counting. Most of the tasks fall under the 1\% line.

\begin{table}[h!]
\caption{Percent of attention scores required to reach 95\% of dense attention performance for different categories of tasks from the RULER benchmark. \kk requirements were measured by performance on contexts from 8,000-128,000 tokens. All samples from Llama-3-8B.}
\label{tab:entropy_correlation}
\centering
\begin{tabular}{lc}
\hline
\textbf{Task Category}           & \textbf{\begin{tabular}[c]{@{}c@{}}\kk Required For\\ 95\% Performance\end{tabular}} \\
Needle In A Haystack             & 0.001\%      \\
Variable Tracking                & 0.11\%       \\
Question Answering               & 0.23\%       \\
Multiple NIAH                    & 0.27\%       \\
Word Counting                    & 8.87\%       \\ 
\hline
\end{tabular}
\end{table}

\paragraph{Layer-Wise Settings for \kk}

Recent work has shown that later layers of a transformer network tend to be less crucial to the computation than earlier ones \citep{gromov2024unreasonableineffectivenessdeeperlayers}. Our method naturally admits a dimension of flexibility as to where efficiency gains are extracted. As some layers may need a larger \texttt{k} budget than others to accurately capture their attention distributions, we explore allowing the value of \kk to vary across layers.

\begin{figure}
    \centering
    \includegraphics[width=0.95\linewidth]{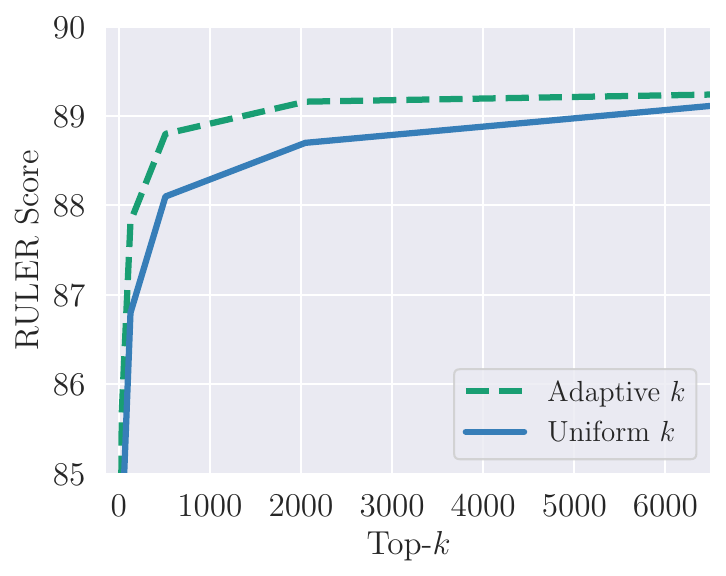}
    \caption{RULER performance of \topk with a fixed vs adaptive \kk budget across layers. The x axis represents the total \kk budget, and lines are given for two \kk budgeting schemes: equal \kk across all layers and a linearly increasing \kk from the first to the last layer.}
    \label{fig:topk_adaptive}
\end{figure}

In \Cref{fig:topk_adaptive} we vary \kk and plot the RULER scores as before, but we allocate our budget for \kk across layers in two different ways. In the uniform strategy, each layer gets the same value of \kk. In the adaptive strategy, we linearly increase \kk from the first to the last layer. In each experiment (point on chart), the total \kk budget remains fixed, e.g. the sum of \kk over all layers is always the same. We find we are able to gain non-trivial performance boosts simply by changing our strategy for how set \kk for each layer given a fixed computational budget.

%% file: Sections/related_work.tex
\section{Related Work}
\label{related_work}

The problem of making transformer model inference more efficient has been studied from many angles. We briefly survey some of the relevant work below.


\subsection{Systems Approaches}
There is a long line of work in systems solutions to scale to long contexts in LLMs at inference. Flash Attention, and later Flash-Attention 2, provide theoretical linear memory complexity over the sequence length \citep{dao2022flashattentionfastmemoryefficientexact, dao2024flashattention}. 
vLLM's implementation of Paged Attention \citep{kwon2023efficient} optimizes for throughput over many requests, and, like Flash Attention, this Paged Attention implements a block matrix multiplication algorithm that allows for memory savings at inference time. vLLM, while very performant, does not perform any offloading and assumes the cache will be able to fit on the GPU.
\citet{liu2023ringattentionblockwisetransformers} propose Ring Attention, which employs blockwise computation for the self-attention and feedforward operations, distributing long sequences over multiple devices and overlapping key-value block communication with the computation of attention. While the method is highly scalable it assumes access to datacenter-level compute. To the best of our knowledge our approach is the first to achieve inference on million token context windows on a single commodity GPU. 

\subsection{Cache Eviction Methods}
A variety of methods exist for evicting tokens from the cache in order to save GPU memory. A straightforward solution is sliding window attention~\citep{longformer}, which keeps only recent tokens of a fixed window size in the cache, and assumes that local interactions dominate. StreamingLLM~\citep{streamingllm} discovered the ``attention-sink" phenomenon and proposed a modified sliding window attention that alleviates the performance degradation in windowed attention. Both these works can fail on long contexts as important tokens in the middle of the context window may become evicted from the cache. Another widely used method is H20 \citep{zhang2023h2oheavyhitteroracleefficient}, a cache eviction strategy that uses information about past tokens to remove tokens from the cache. While effective in increasing decode speeds, this method is not dynamically adjustable to different queries, leading some tokens to be evicted that may be important later. Our method keeps all tokens in the cache offloaded to cheap and plentiful CPU memory, but accelerates the lookup process with fast approximate nearest neighbors.

\subsection{\Topk and Dynamic Algorithms}
The closest approaches to our method in the literature are all based around the initial work on \topk in \citet{guptaMemoryefficientTransformersTopk2021}. This method computes all scores directly before filtering out the \topk, incurring quadratic computation and demanding the entire KV cache fits on GPU memory. \citet{chen2024magicpiglshsamplingefficient} also accelerate decoding by selecting relevant tokens, but take an approach based on sampling attention distributions. 
\citet{singhania2024lokilowrankkeysefficient} construct a low-cost proxy to full attention using PCA, which informs the token subset for the attention computation. 
However, these solutions do not support context lengths past a few hundred thousand.

\citet{tang2024questqueryawaresparsityefficient} dynamically select relevant tokens from the cache during decoding, but their method relies on a heuristic approximation of \topk divided over cache pages, whereas our method utilizes a nearest-neighbor data structure.
 The concurrent works of \citep{liu2024retrievalattentionacceleratinglongcontextllm} and \citep{zhang2024pqcacheproductquantizationbasedkvcache} bear a resemblance to our method but enforce exactly which kind of nearest-neighbor search algorithm is employed: a custom database in the first and the method of PQ quantization in the second. Our method both generalizes their method to allow for any database and is extended to 1 million length contexts and beyond.






%% file: Sections/conclusion.tex
\section{Conclusion}
\label{conclusion}

In this work we demonstrate the capability of a \topk attention mechanism to operate at the million token scale on a single GPU. We achieve sublinear complexity and evaluate at over 95\% of dense attention accuracy on common benchmarks while using only 2\% of the context length on average in the attention block. This exploitation of attention sparsity opens up new directions for efficient and viable solutions to long context inference in language models. Our investigation of attention distributions across layers points to future variations of our method that smoothly adapt the \topk method across tasks and layers of a given model suggesting our \topk approach can be used to achieve optimal compute-performance tradeoffs given a specific deployment context.

%% file: Sections/appendix.tex
\newpage
\appendix
\onecolumn
\section{Appendix}
\label{appendix}

%

\subsection{Distribution of Attention Scores}
\label{app:attention_distribution}
In general, 1\% of the total attention scores are sufficient for providing 95\% of the performance of dense attention on Open LLM Leaderboard, AlpacaEval, and RULER. However, within the subtasks for a given benchmark there is variation in this \kk-required threshold. This variation is highly correlated with the a measurement we call the \textit{attention entropy}. Attention entropy is calculated by taking the Shannon entropy of a single row of an attention matrix after the softmax transformation is applied, and averaging that over multiple tokens of generation on many different samples of text. 

When treating a single, soft-maxed row of an attention matrix as a probability distribution, entropy serves as a good descriptor of how concentrated, or "sparse" it is. The entropy of a maximally concentrated attention distribution is zero, while completely uniform attention scores would have an entropy of the logarithm total scores. Thus low entropy indicates sparse, or concentrated attention. In \Cref{tab:entropy_correlation_app} we show the attention entropy calculated from the first ten tokens of generation from fifty samples of text in each task category. The attention entropy values in the table have a Pearson correlation coefficient of 0.85 with the \kk-required thresholds for 95\% performance in those tasks. 

\begin{figure}[h]
    \centering
    \includegraphics[width=0.76\linewidth]{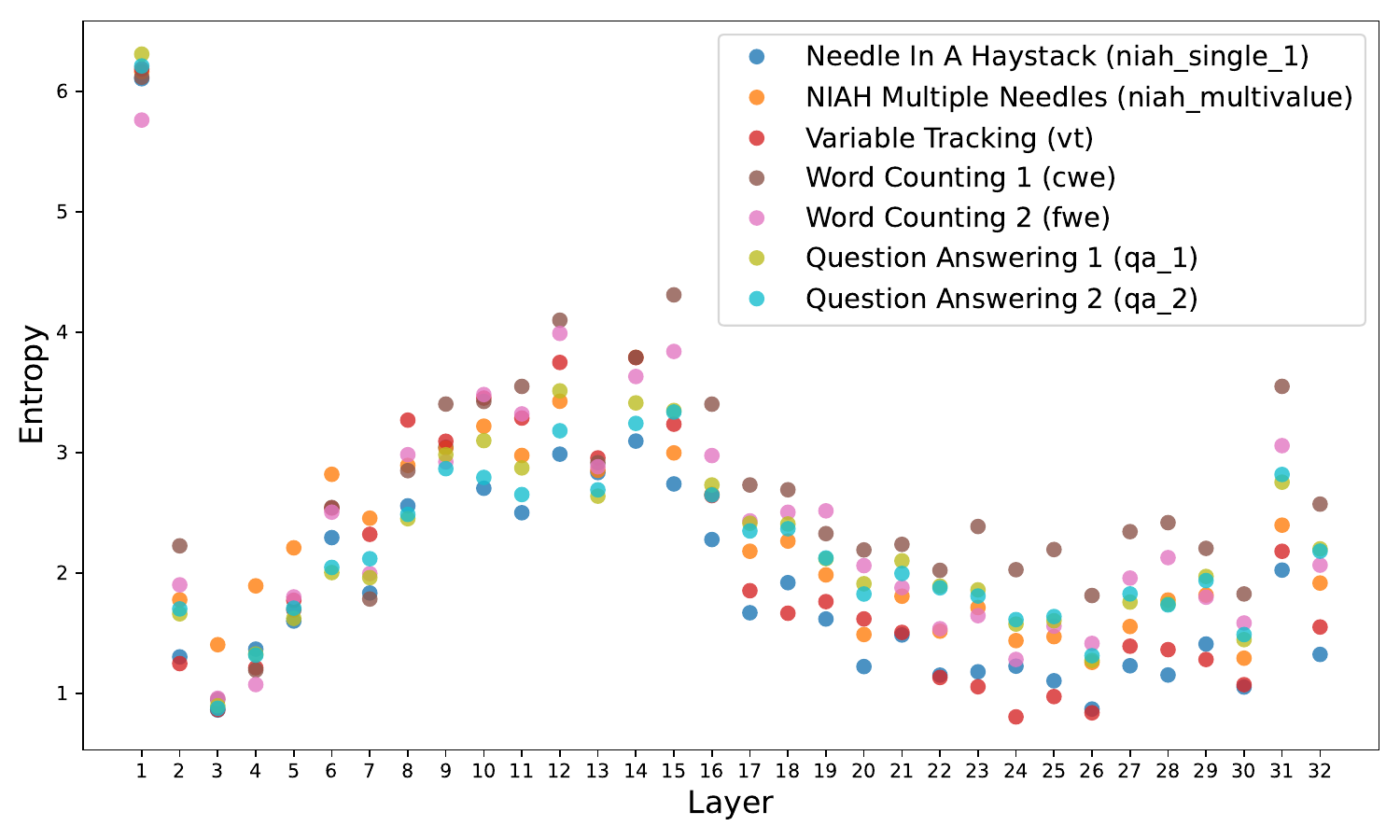}
    \caption{Attention entropy by layer, colored by task category.}
    \label{fig:entropy_tasks_app}
\end{figure}

In addition to looking at the attention distributions across tasks, we investigate if there are any systematic trends in the attention sparsity across layers of a model. \Cref{fig:entropy_tasks_app} shows the attention entropy for RULER subtasks when plotted by layer, and \Cref{fig:75_mass_hist} shows the number of scores required to cover 75\% of the full attention distribution, with a histogram plotting this value over hundreds of samples of Wikipedia text. Note that in both figures, the first layer clearly stands out as having the least concentrated attention distributions.

\begin{table}[h]
\centering
\caption{Percent of attention scores required to reach 95\% of dense attention performance for different categories of tasks, along with the average entropy of the attention vectors for tokens generated from those tasks.}
\label{tab:entropy_correlation_app}
\begin{tabular}{lcc}
\hline
\textbf{Task Category}     & \textbf{\begin{tabular}[c]{@{}c@{}}\kk Required For\\ 95\% Performance\end{tabular}} & \textbf{\begin{tabular}[c]{@{}c@{}}Attention\\ Entropy\end{tabular}} \\ \hline
Needle In A Haystack       & 0.001\%                   & 1.93 \\
Variable Tracking          & 0.11\%                    & 2.11 \\
Question Answering         & 0.23\%                    & 2.27 \\
Multiple NIAH              & 0.27\%                    & 2.33 \\
Word Counting              & 8.87\%                    & 2.68 \\ \hline
\textbf{\kk-Required - Entropy Correlation ($r$)} & \multicolumn{2}{c}{0.847}        \\ \hline
\end{tabular}
\end{table}

\input{Sections/32_histogram_figure}

\subsection{Additional RULER and AlpacaEval Results}

\begin{figure}[htbp]
    \centering
    \begin{subfigure}{0.3\textwidth}
        \includegraphics[width=\linewidth]{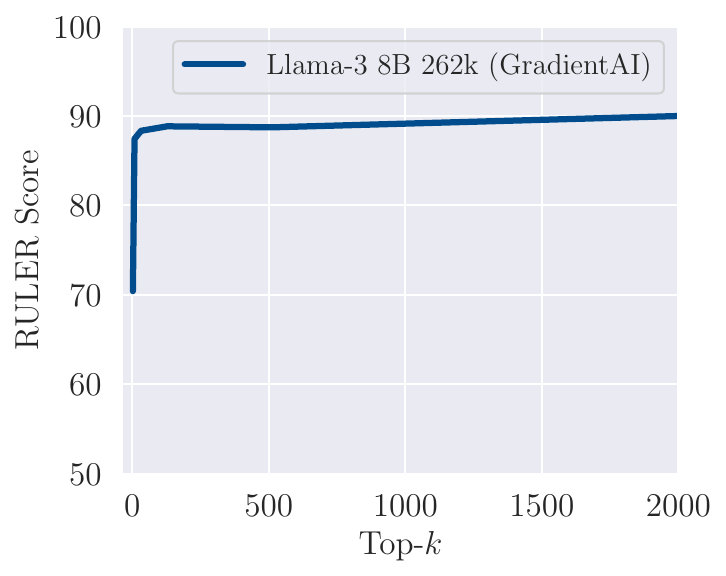}
        \caption{Context length 4096}
    \end{subfigure}
    \hfill
    \begin{subfigure}{0.3\textwidth}
        \includegraphics[width=\linewidth]{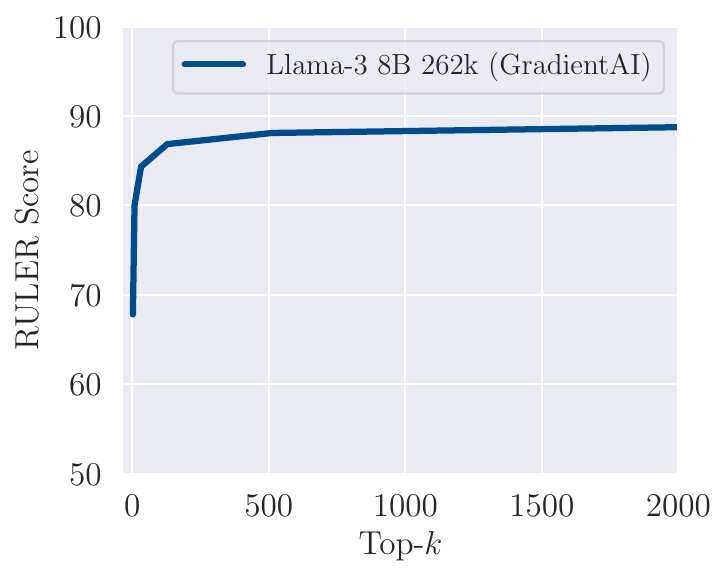}
        \caption{Context length 8192}
    \end{subfigure}
    \hfill
    \begin{subfigure}{0.3\textwidth}
        \includegraphics[width=\linewidth]{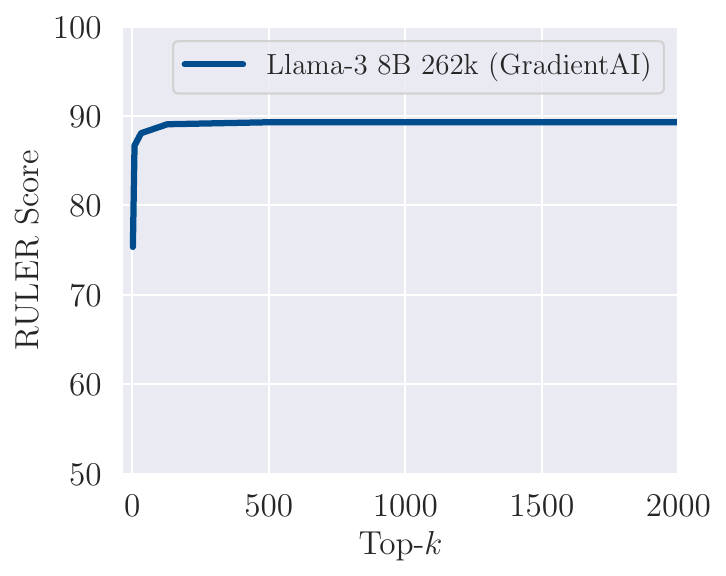}
        \caption{Context length 16384}
    \end{subfigure}

    \vspace{0.5cm} 

    \begin{subfigure}{0.3\textwidth}
        \includegraphics[width=\linewidth]{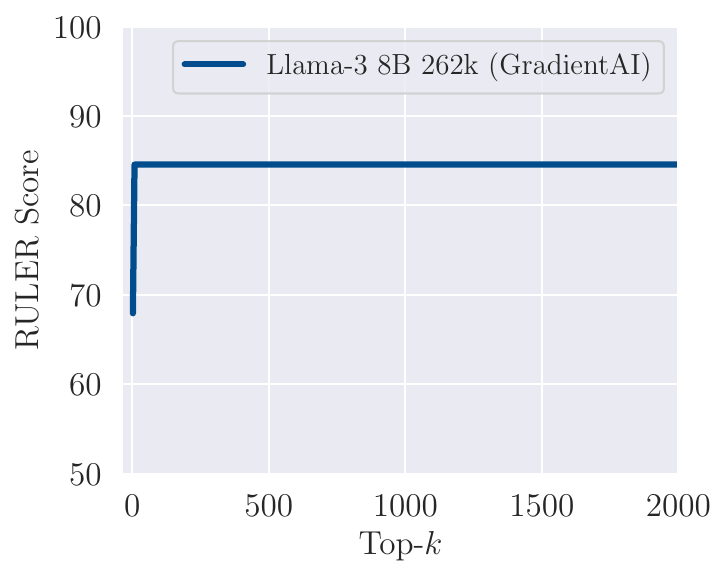}
        \caption{Context length 32768}
    \end{subfigure}
    \hfill
    \begin{subfigure}{0.3\textwidth}
        \includegraphics[width=\linewidth]{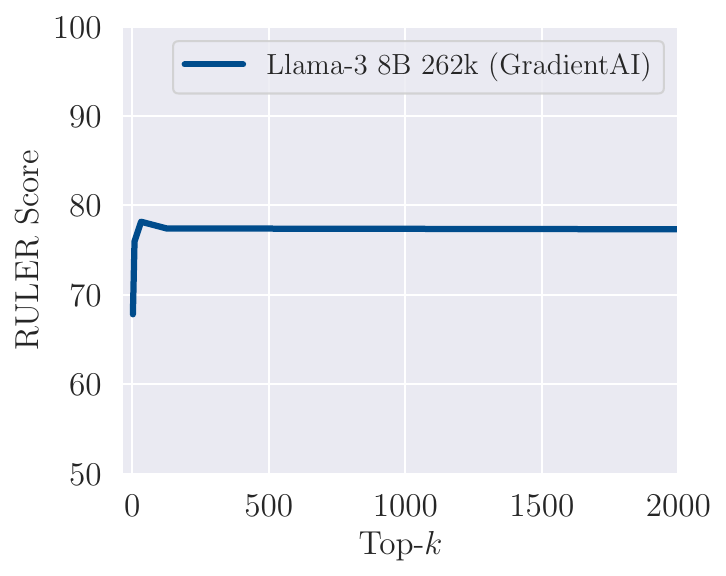}
        \caption{Context length 65536}
    \end{subfigure}
    \hfill
    \begin{subfigure}{0.3\textwidth}
        \includegraphics[width=\linewidth]{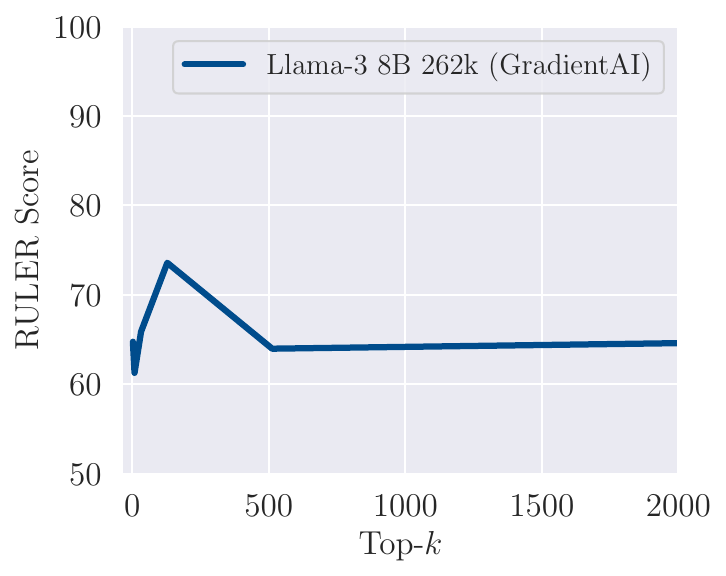}
        \caption{Context length 131072}
    \end{subfigure}

    \caption{Results for RULER over various context lengths. This shows the same behavior as Open LLM Leaderboard and AlpacaEval where very few attention scores (very low \kk) are sufficient to achieve near dense attention performance.}
\end{figure}

\begin{figure}[htbp]
    \centering
    \includegraphics[width=0.35\linewidth]{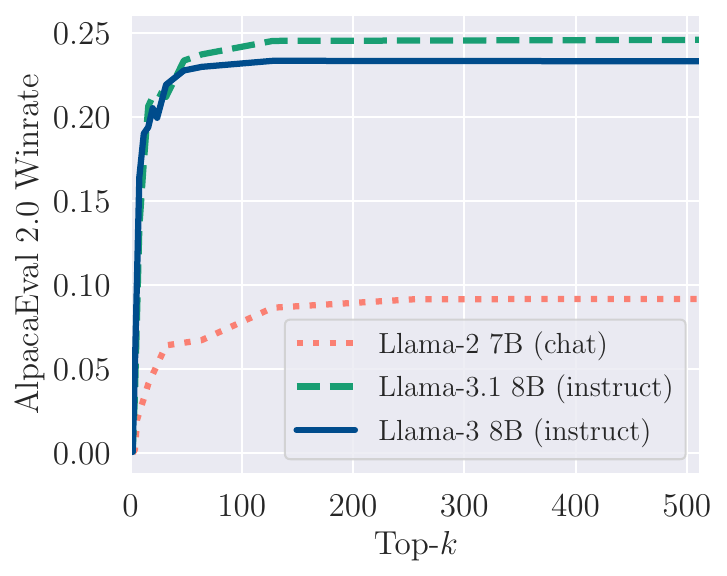}
    \includegraphics[width=0.35\linewidth]{Figures/TopK_Alpaca_Tasks_Size.pdf}
    \caption{AlpacaEval 2.0 results for various models. Left compares different generations of Llama instruction tuned models. Right investigates how models of different sizes handle small values of \kk.}
    \label{fig:alpaca_eval_topk_results_app}
\end{figure}

%% file: Sections/32_histogram_figure.tex
\begin{figure}
    \centering
    \includegraphics[width=0.20\linewidth]{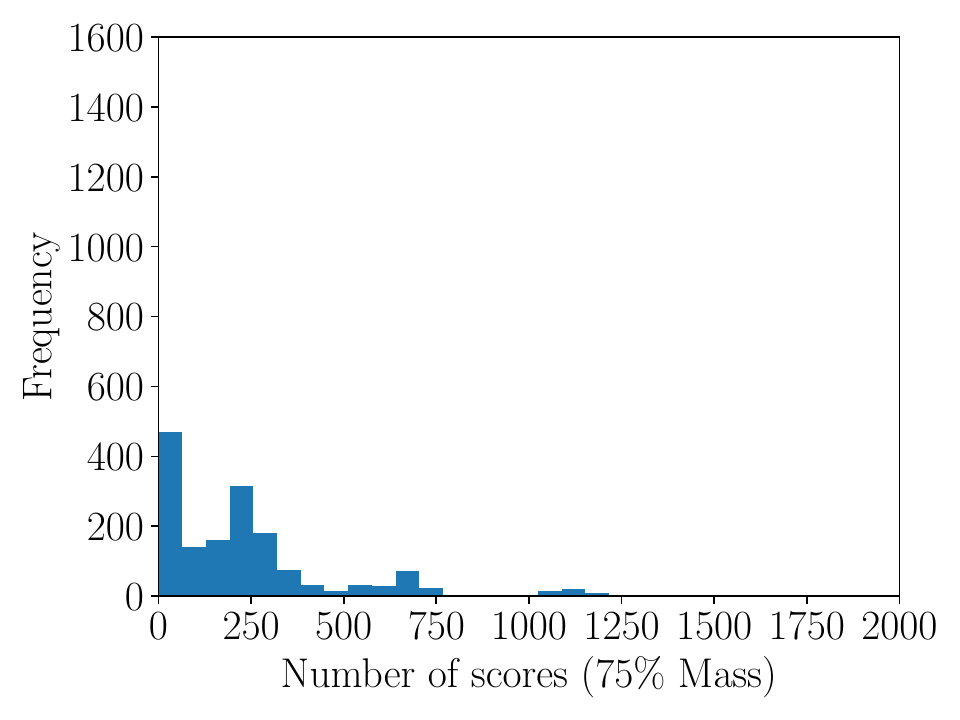}
    \includegraphics[width=0.20\linewidth]{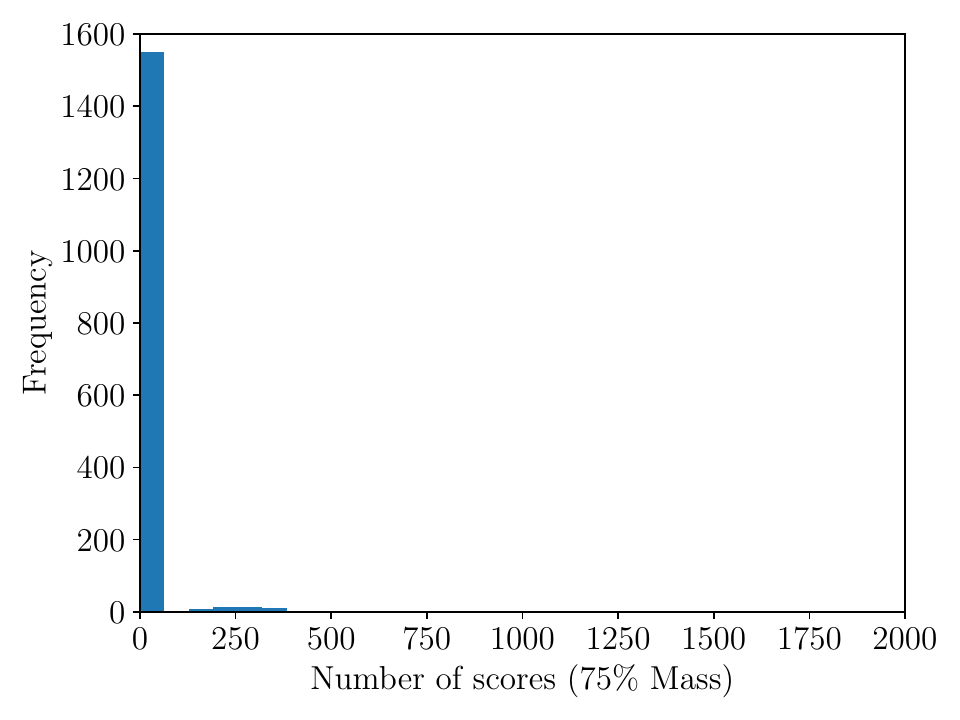}
    \includegraphics[width=0.20\linewidth]{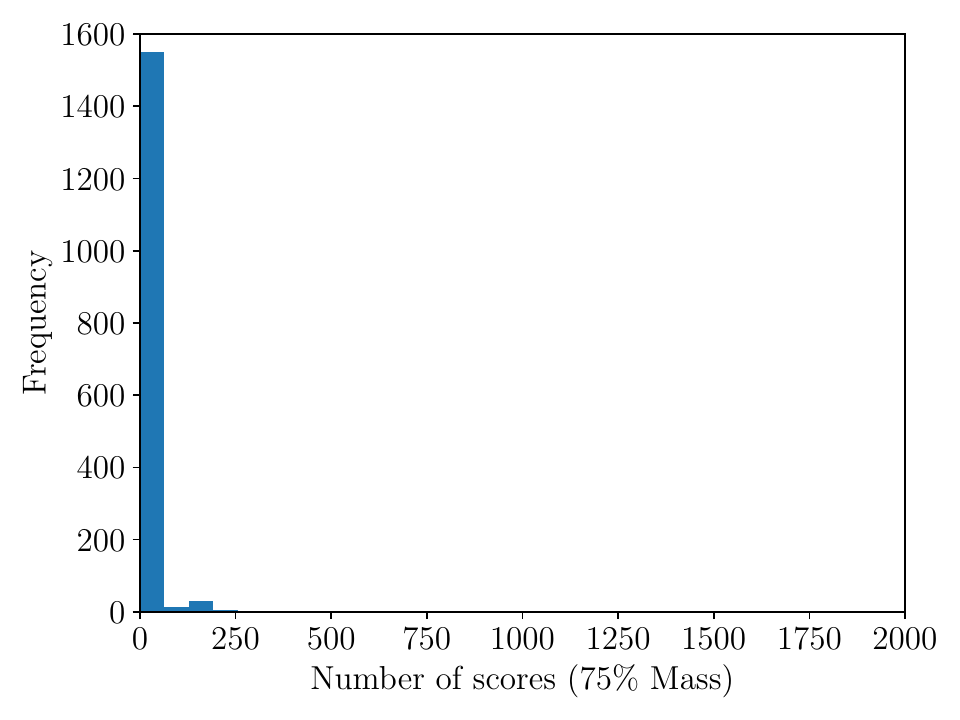}
    \includegraphics[width=0.20\linewidth]{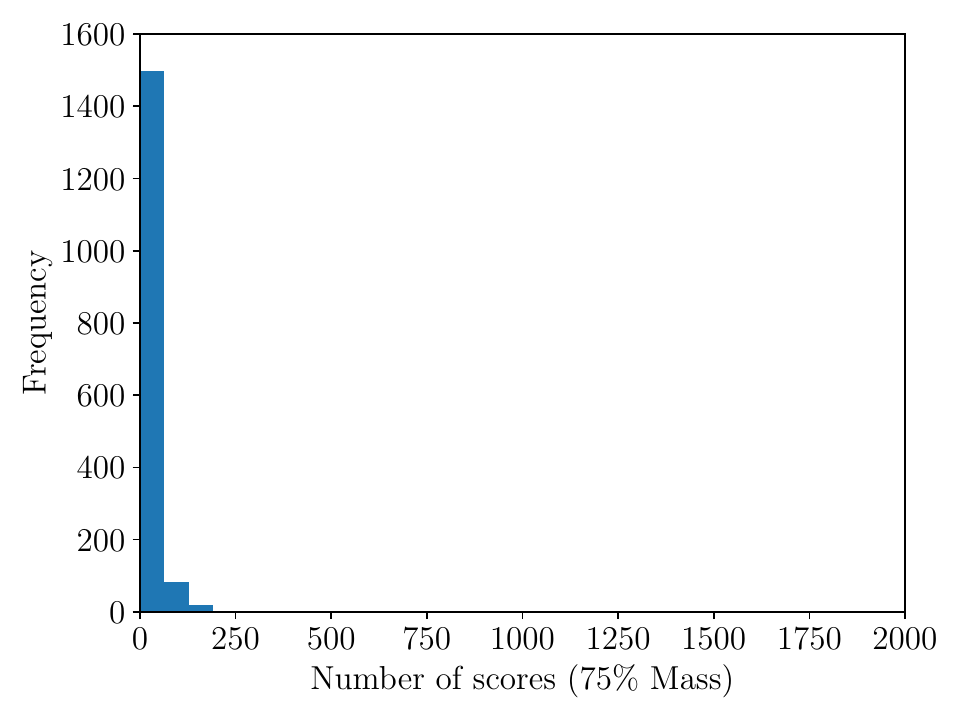}
    \includegraphics[width=0.20\linewidth]{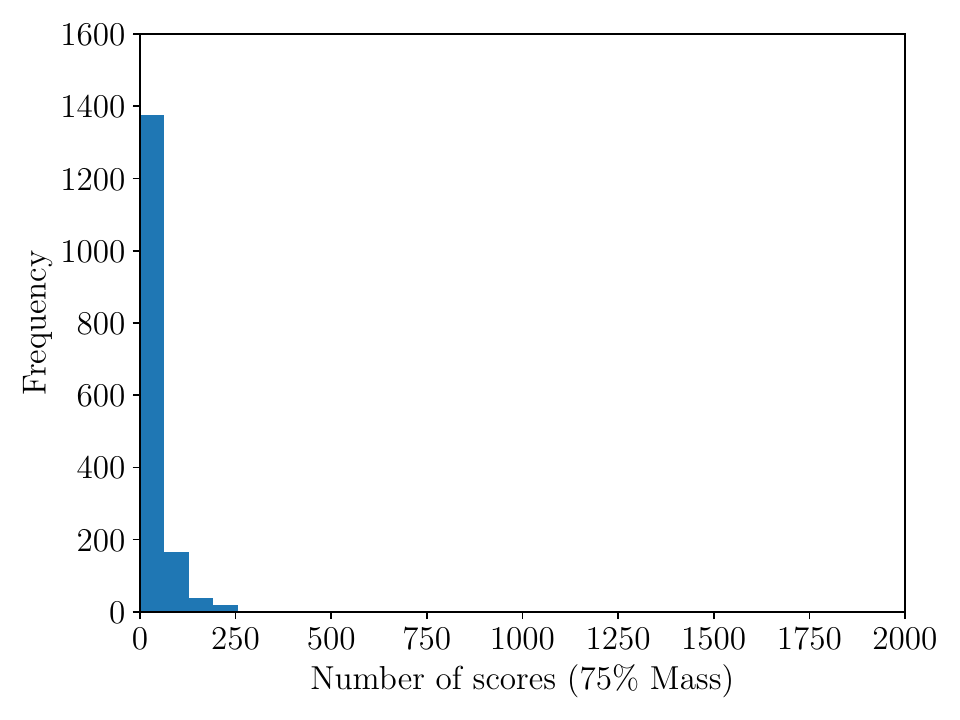}
    \includegraphics[width=0.20\linewidth]{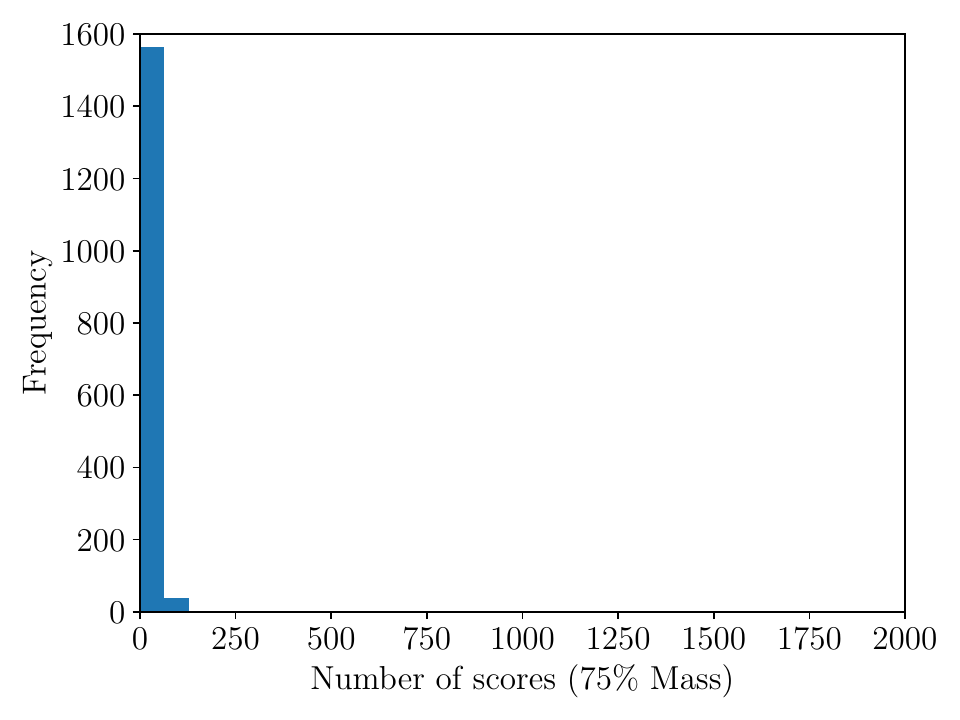}
    \includegraphics[width=0.20\linewidth]{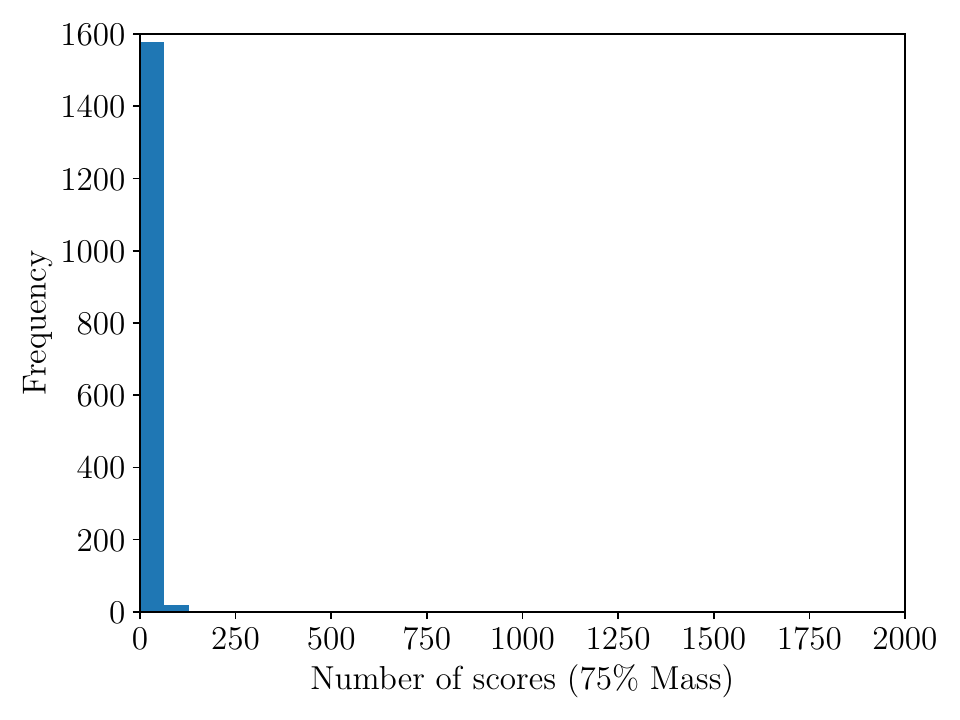}
    \includegraphics[width=0.20\linewidth]{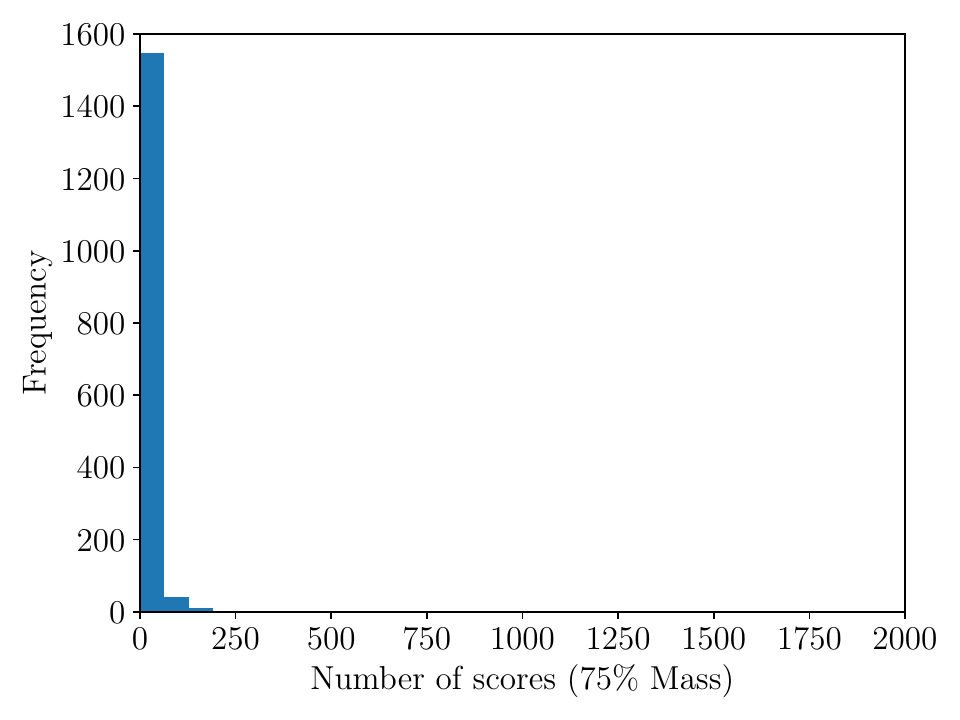}
    \includegraphics[width=0.20\linewidth]{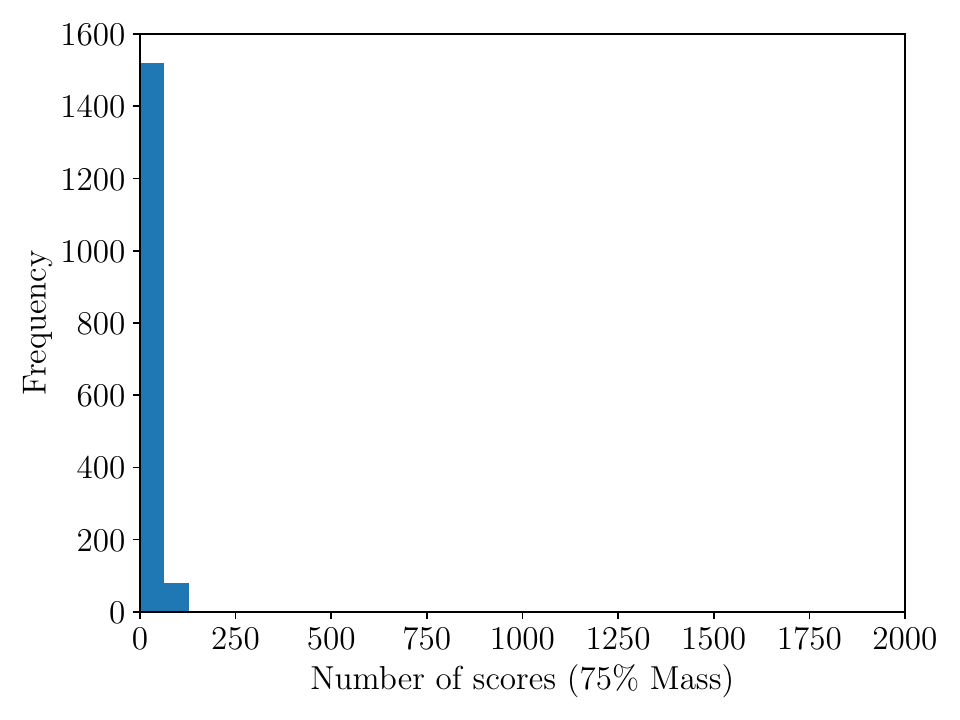}
    \includegraphics[width=0.20\linewidth]{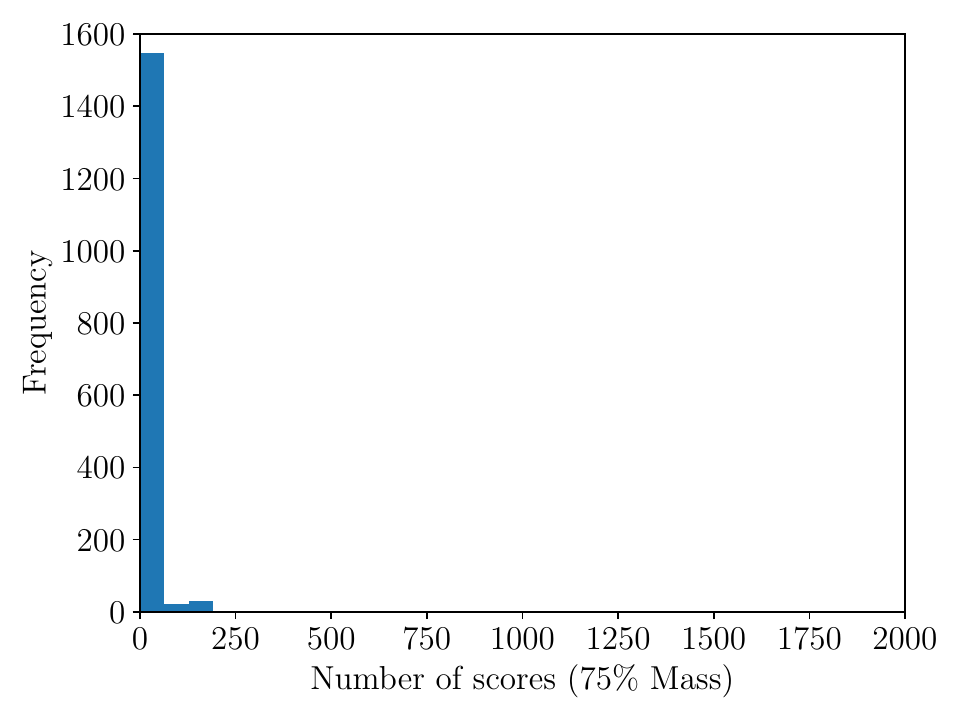}
    \includegraphics[width=0.20\linewidth]{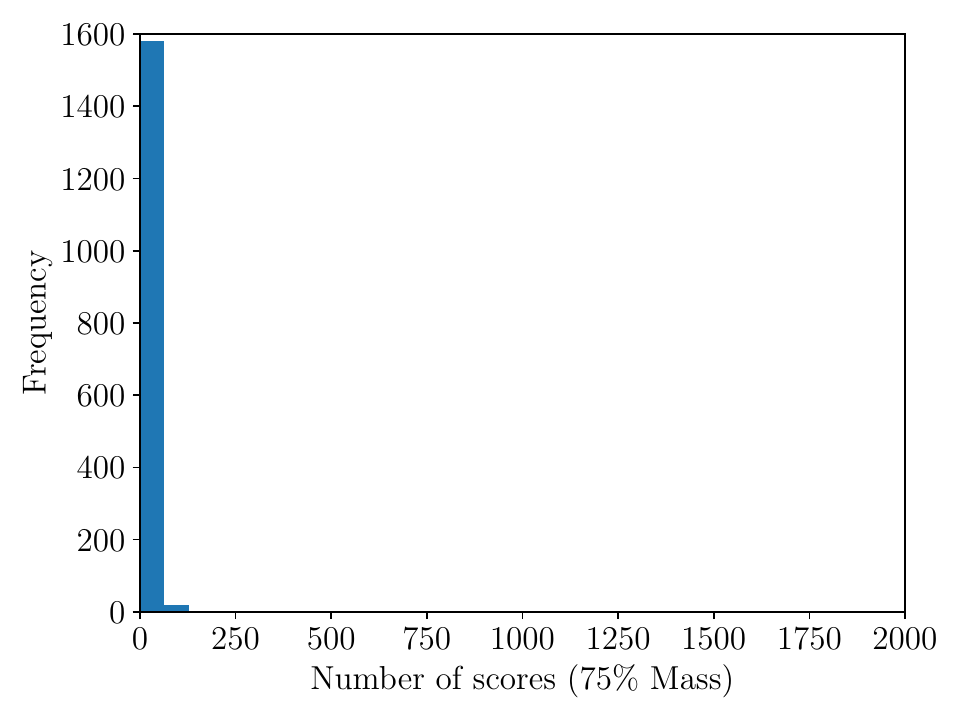}
    \includegraphics[width=0.20\linewidth]{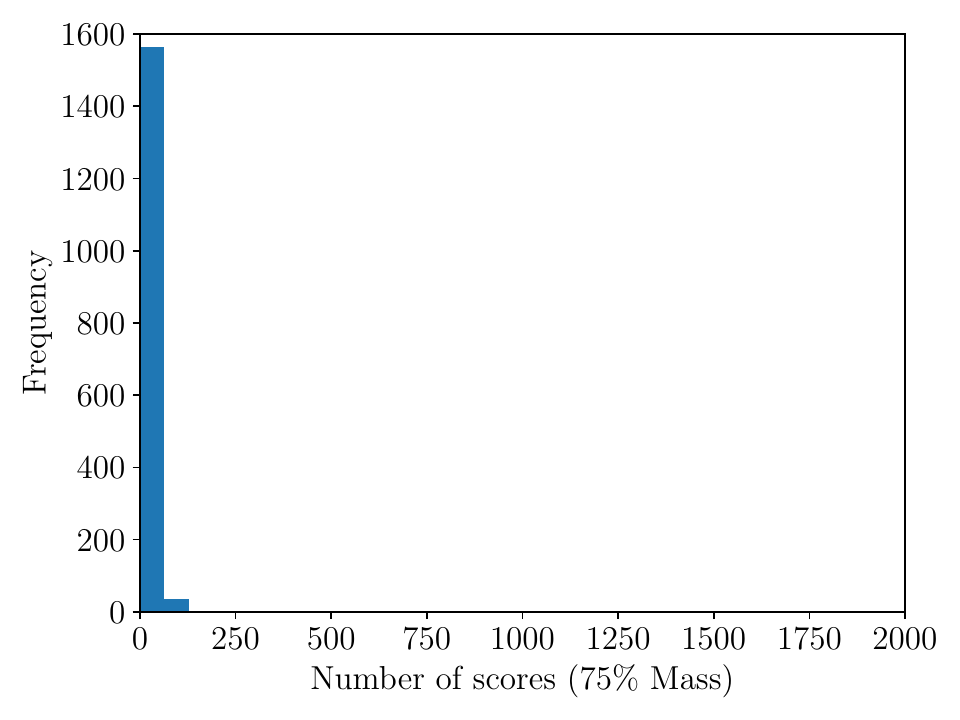}
    \includegraphics[width=0.20\linewidth]{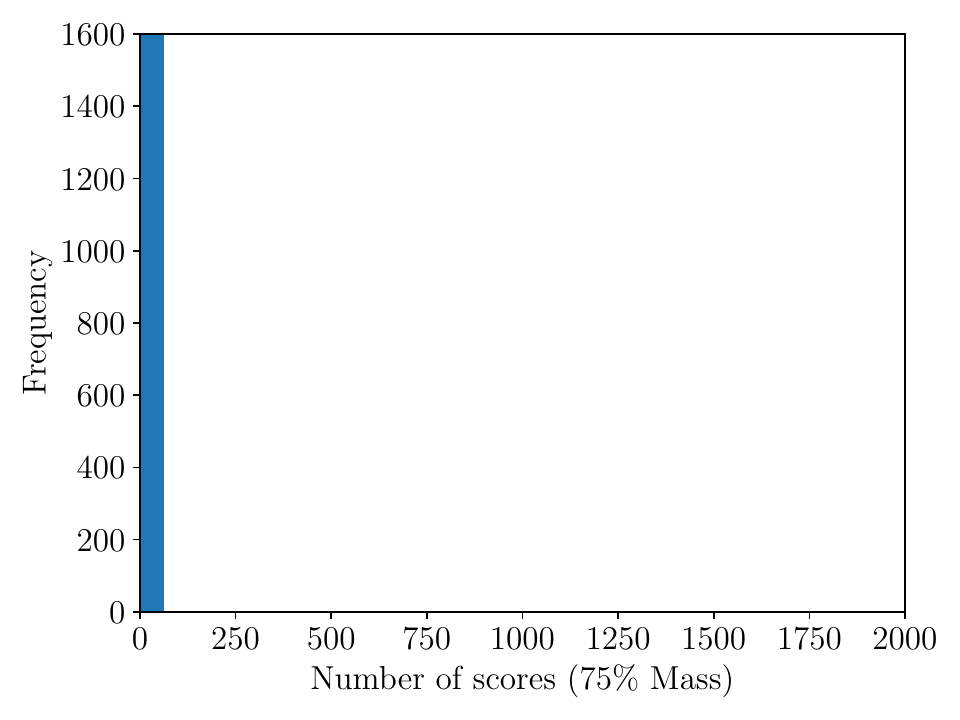}
    \includegraphics[width=0.20\linewidth]{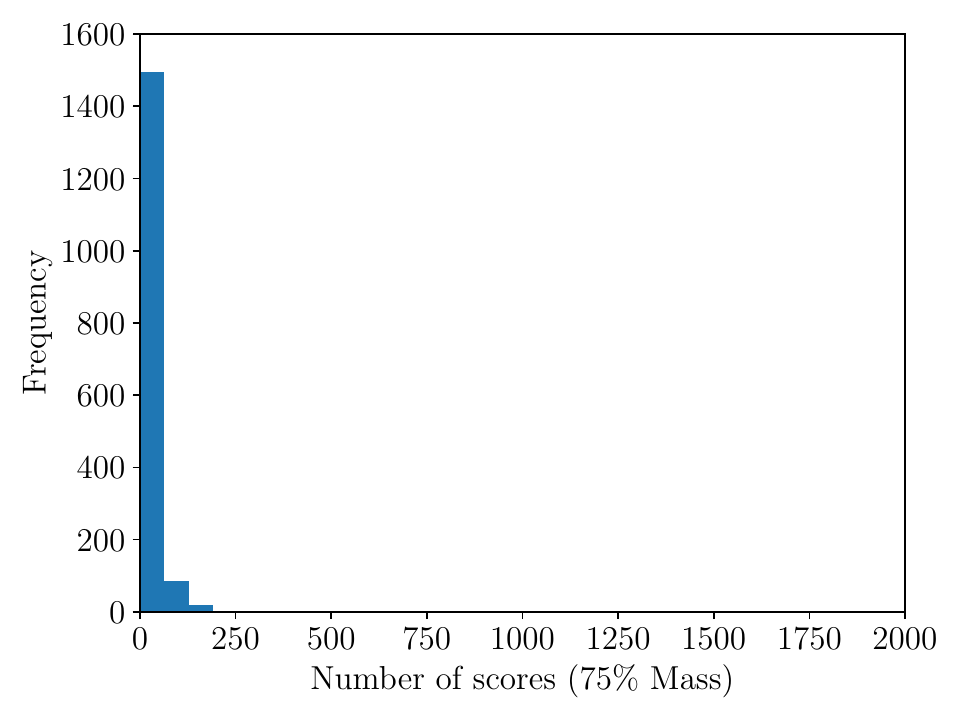}
    \includegraphics[width=0.20\linewidth]{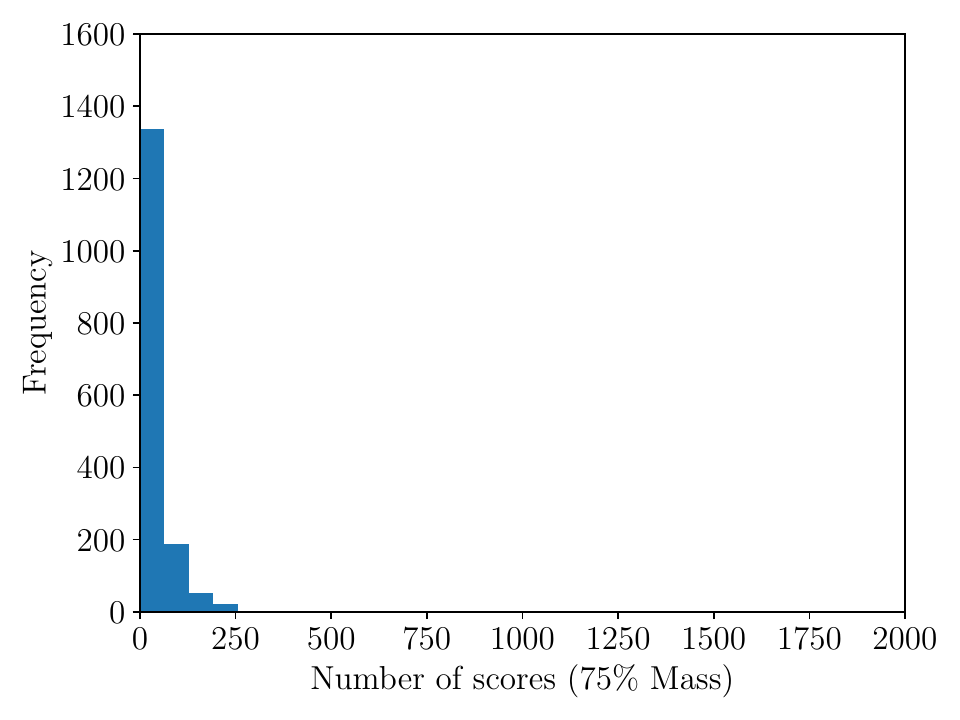}
    \includegraphics[width=0.20\linewidth]{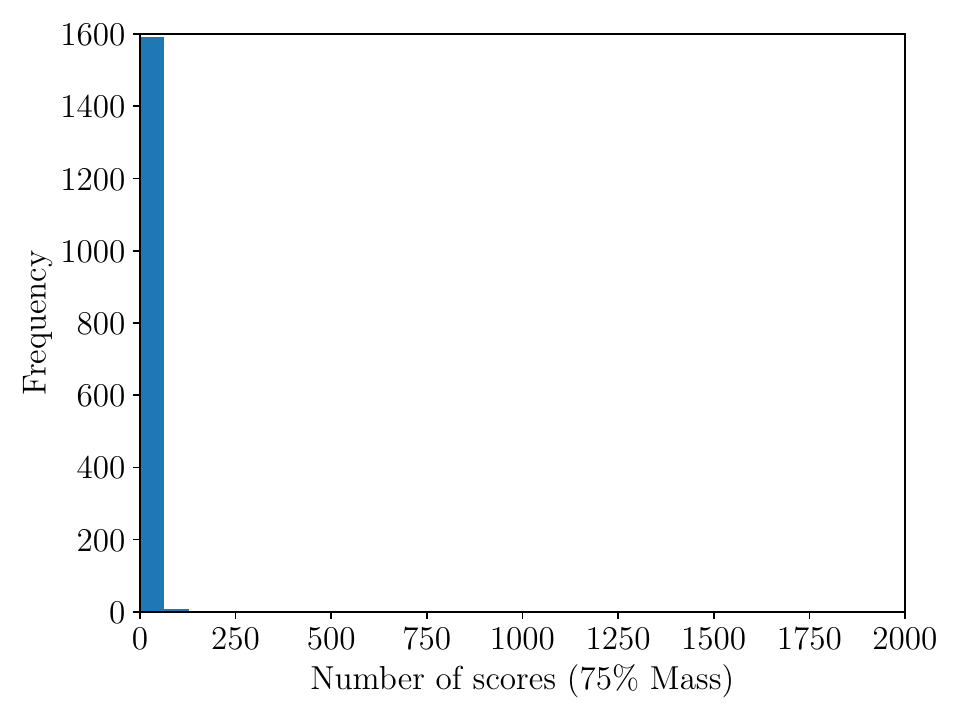}
    \includegraphics[width=0.20\linewidth]{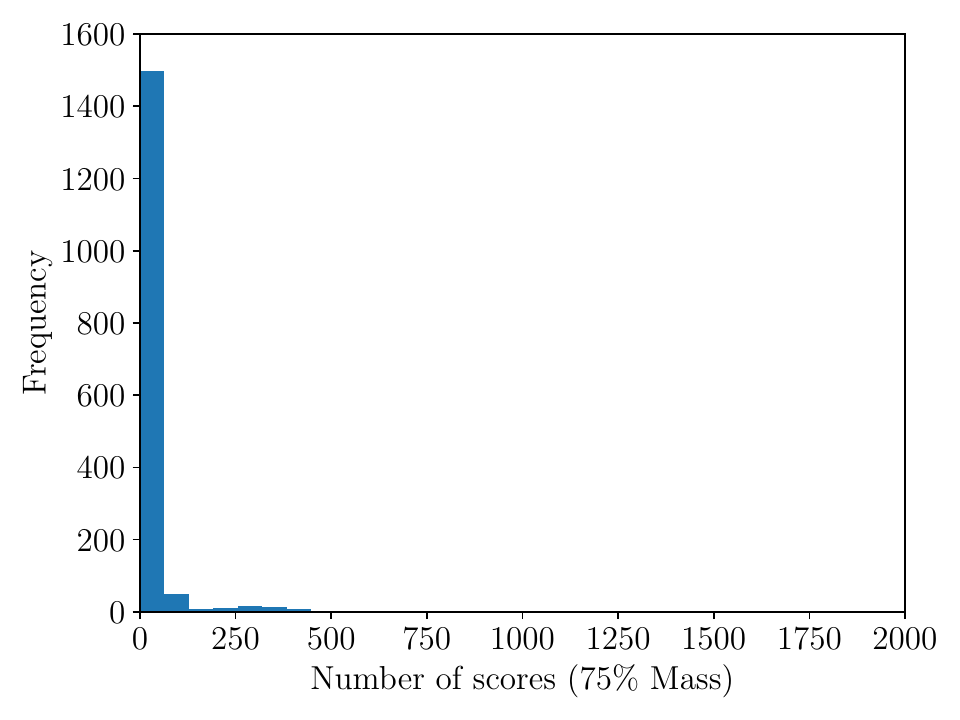}
    \includegraphics[width=0.20\linewidth]{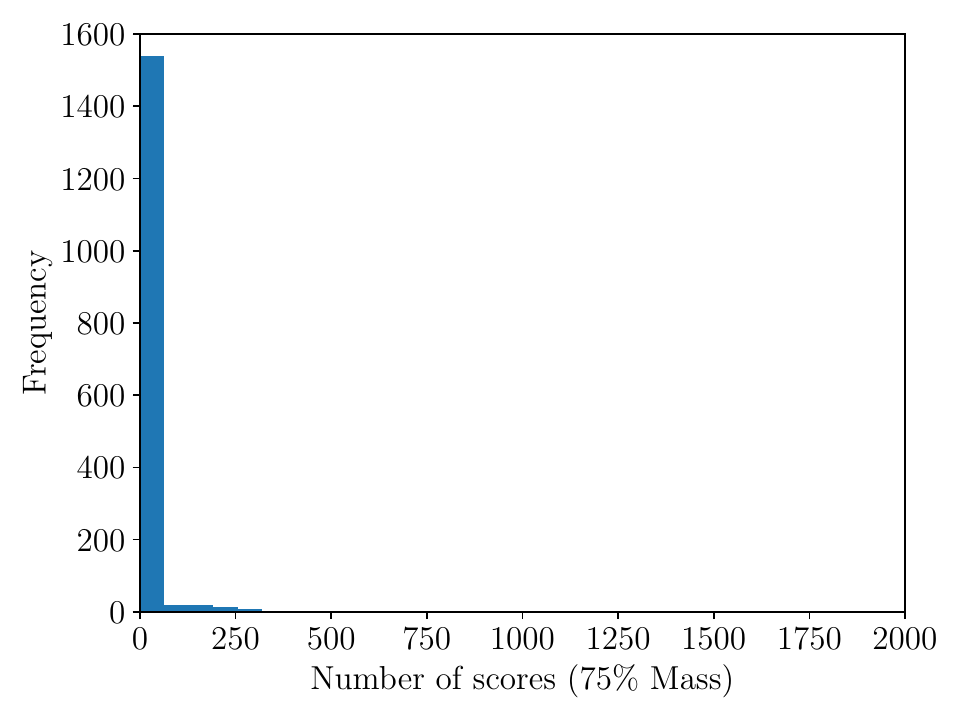}
    \includegraphics[width=0.20\linewidth]{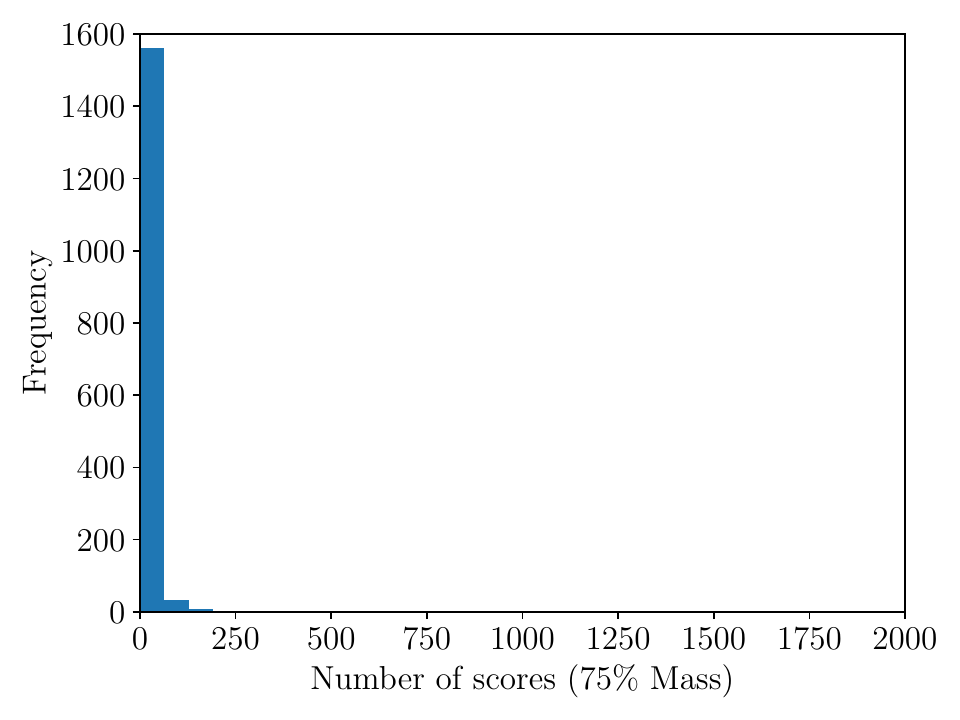}
    \includegraphics[width=0.20\linewidth]{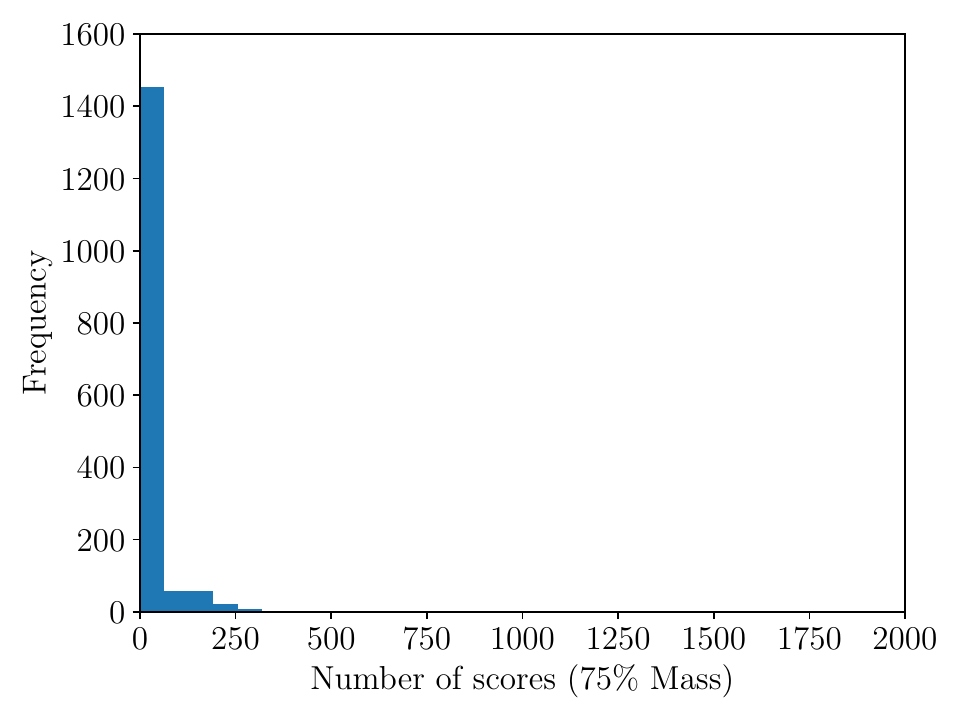}
    \includegraphics[width=0.20\linewidth]{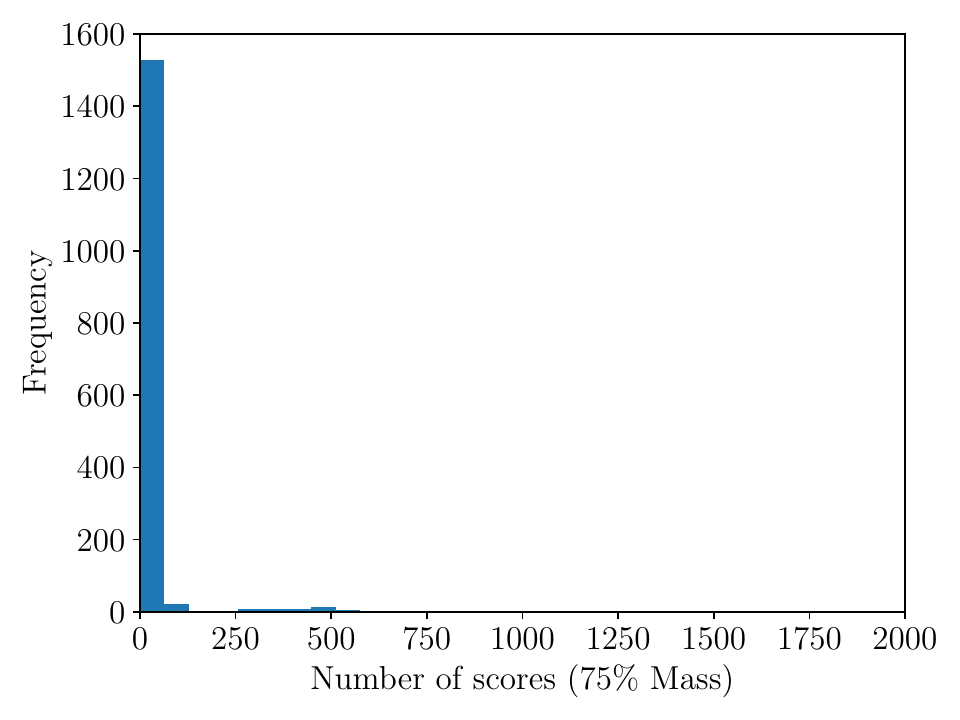}
    \includegraphics[width=0.20\linewidth]{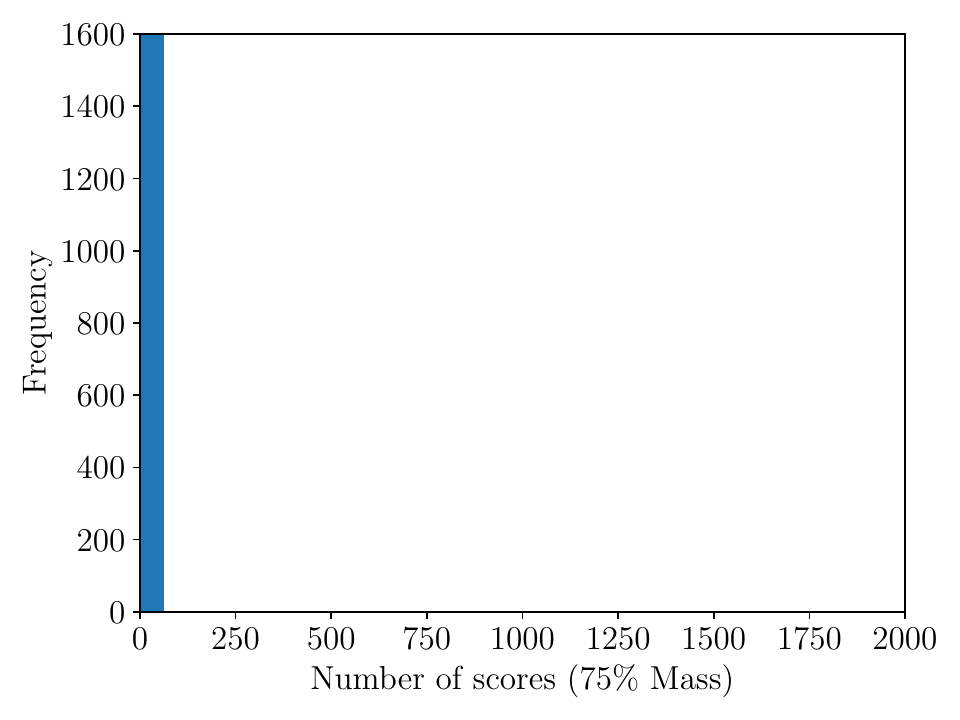}
    \includegraphics[width=0.20\linewidth]{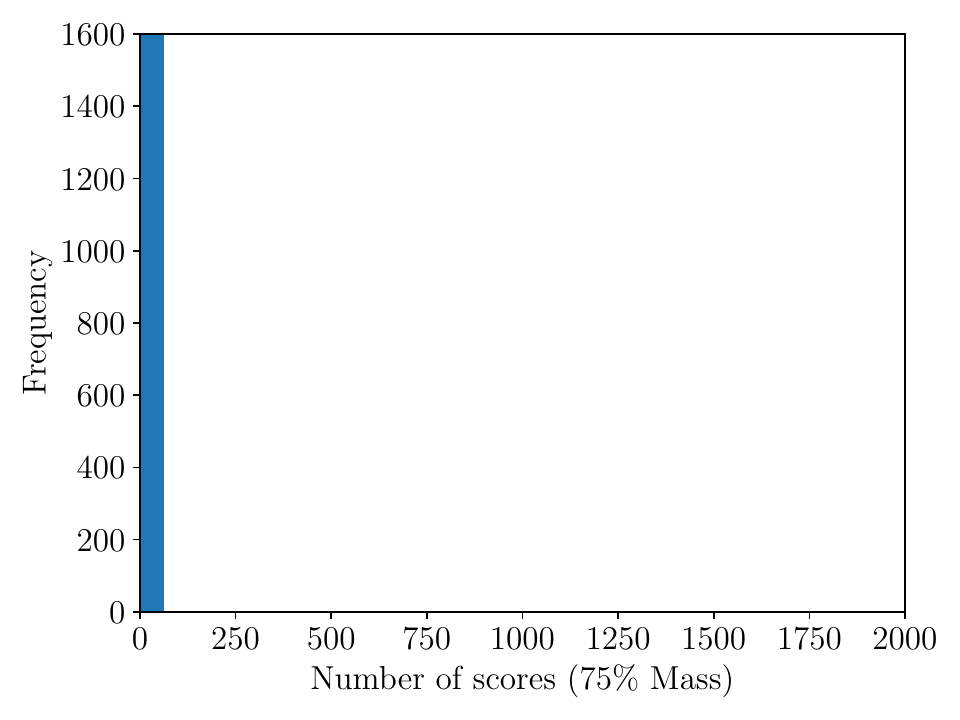}
    \includegraphics[width=0.20\linewidth]{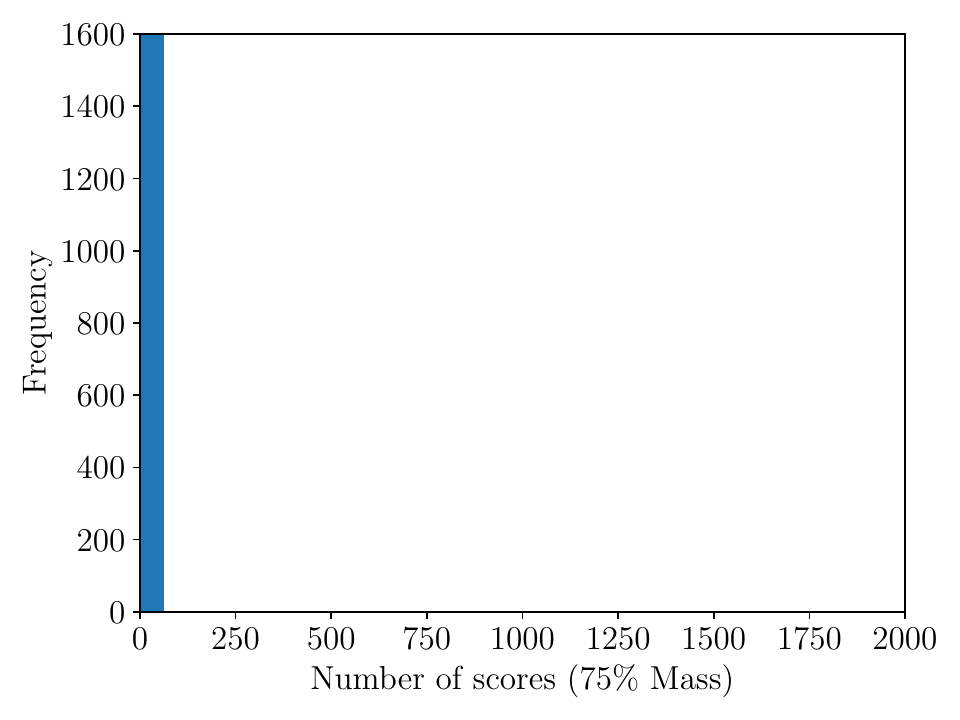}
    \includegraphics[width=0.20\linewidth]{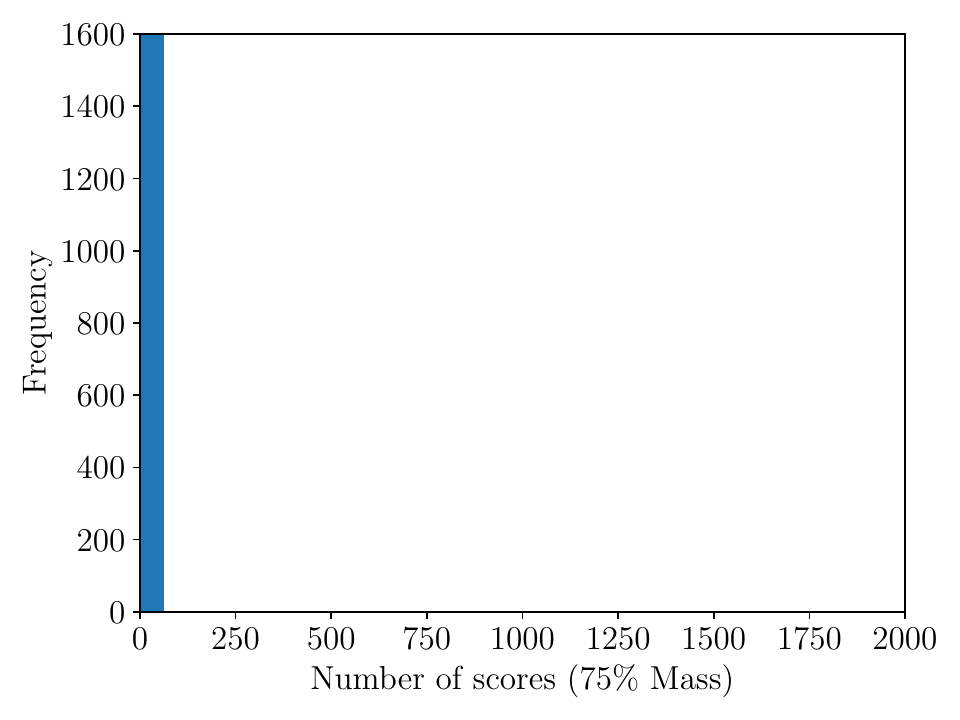}
    \includegraphics[width=0.20\linewidth]{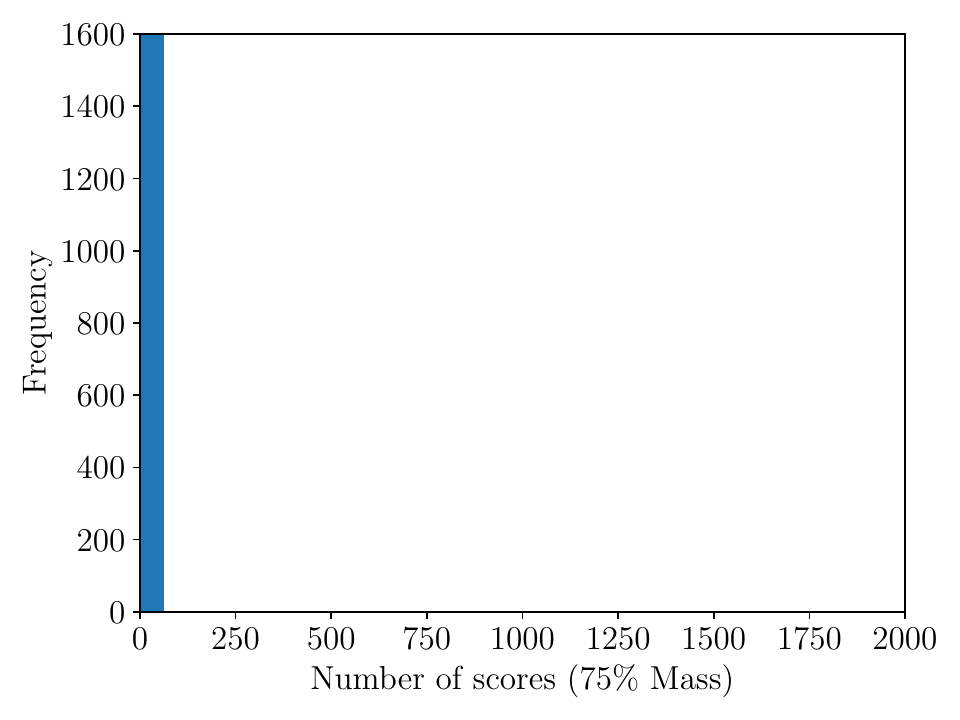}
    \includegraphics[width=0.20\linewidth]{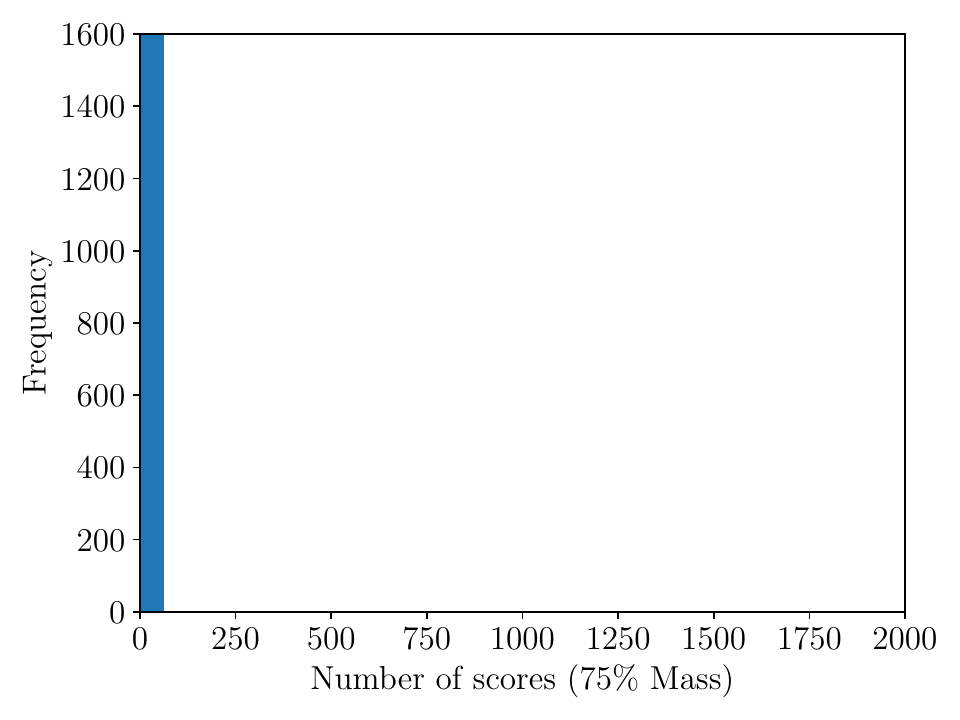}
    \includegraphics[width=0.20\linewidth]{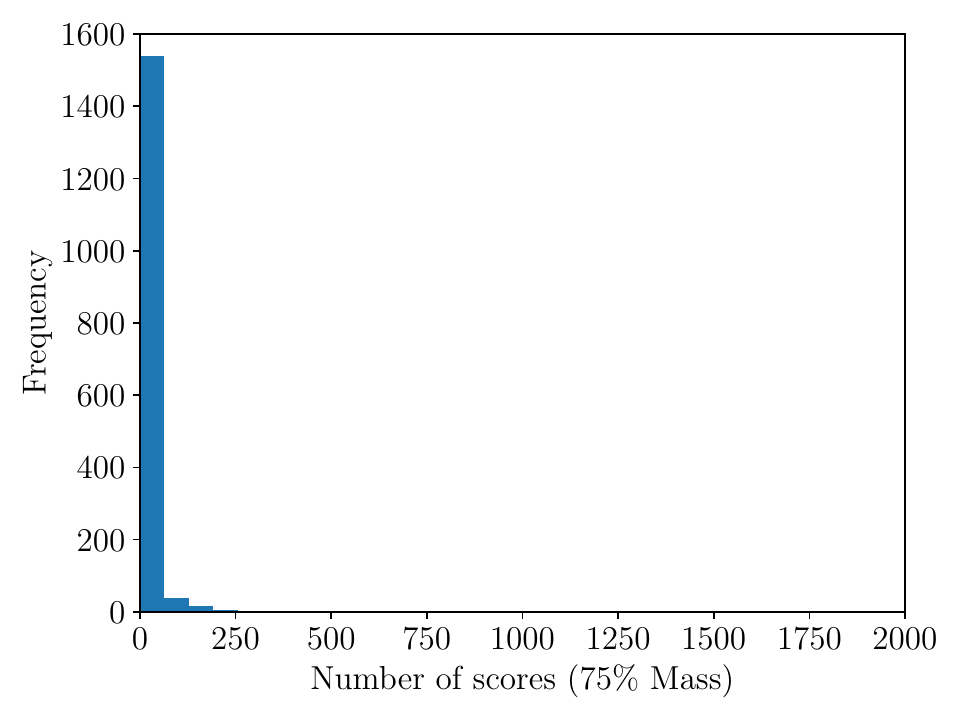}
    \includegraphics[width=0.20\linewidth]{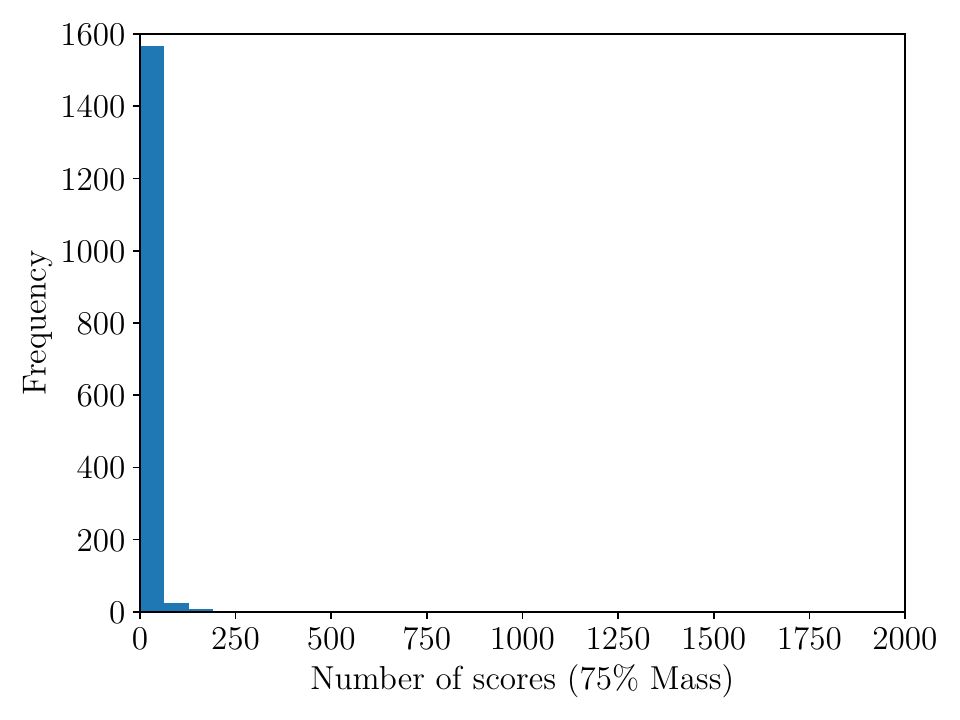}
    \includegraphics[width=0.20\linewidth]{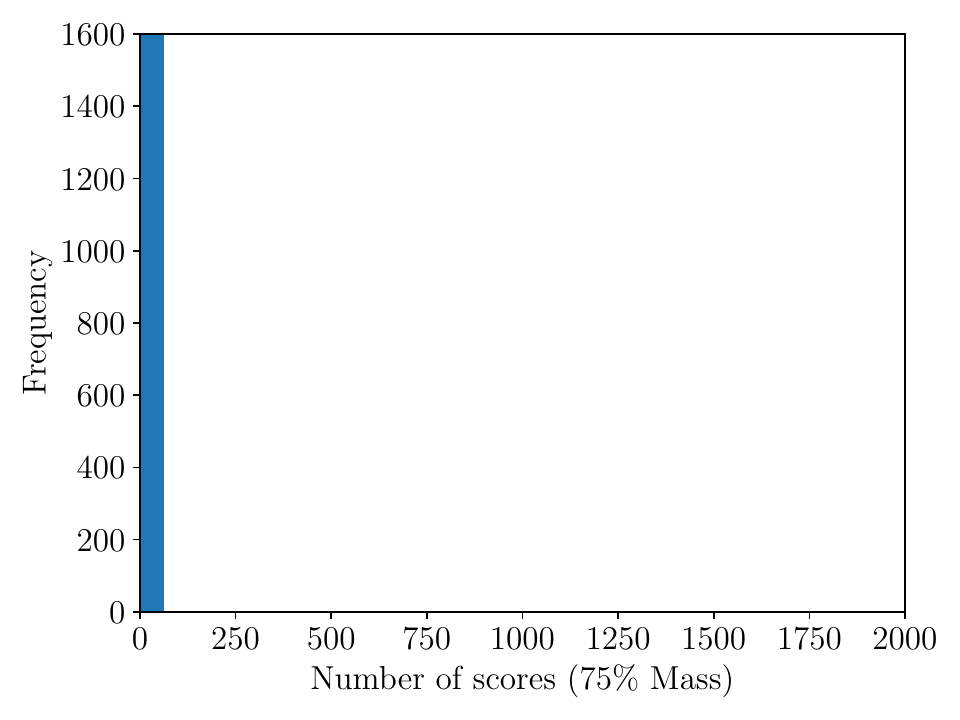}
    \includegraphics[width=0.20\linewidth]{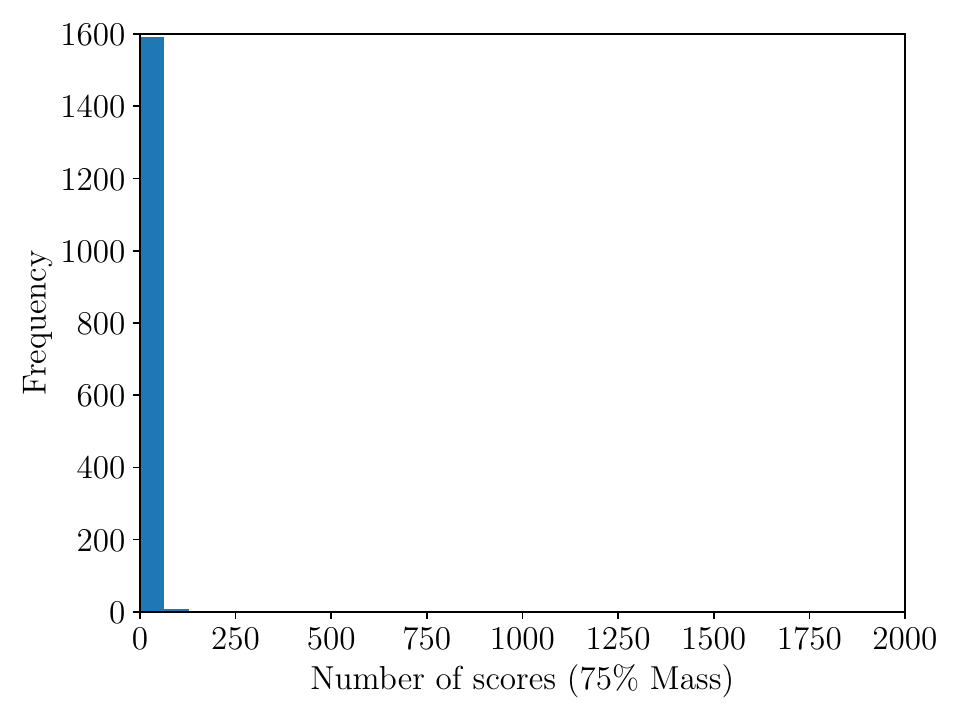}
    \includegraphics[width=0.20\linewidth]{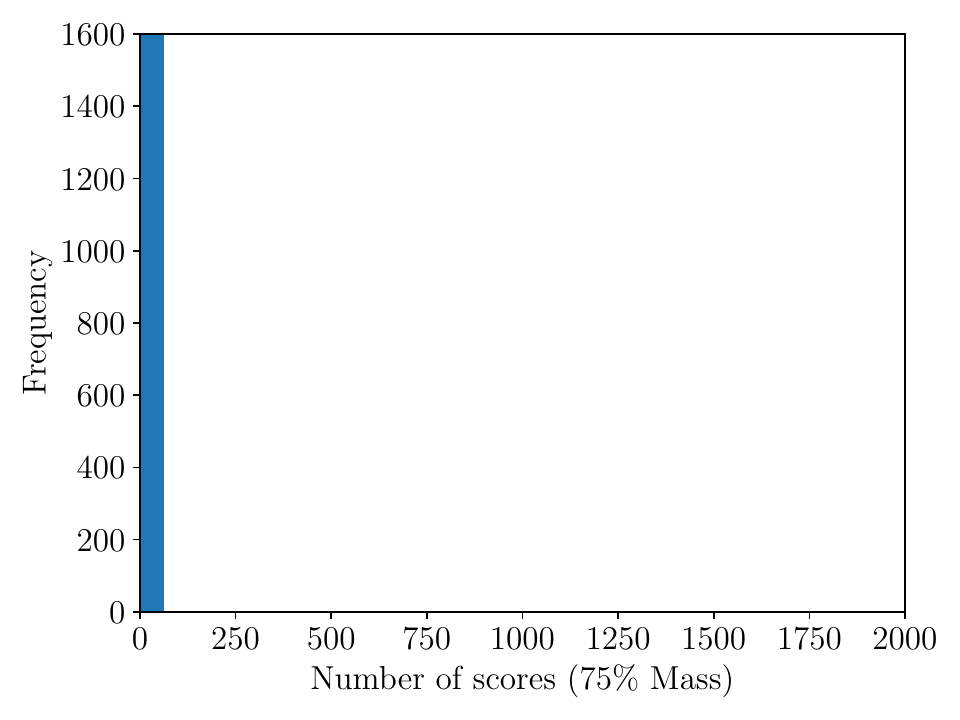}
    \caption{All $32$ layers are plotted in order, where the top row represents layers 1, 2, 3, and 4 and last row represents layers 29, 30, 31, and 32.}
    \label{fig:enter-label}
\end{figure}

%% file: topk_paper.bbl
\begin{thebibliography}{41}
\providecommand{\natexlab}[1]{#1}
\providecommand{\url}[1]{\texttt{#1}}
\expandafter\ifx\csname urlstyle\endcsname\relax
  \providecommand{\doi}[1]{doi: #1}\else
  \providecommand{\doi}{doi: \begingroup \urlstyle{rm}\Url}\fi

\bibitem[AI(2024)]{Meta2024llama3_2}
AI, M.
\newblock Llama 3.2: Revolutionizing edge ai and vision with open, customizable models.
\newblock \url{https://ai.meta.com/blog/llama-3-2-connect-2024-vision-edge-mobile-devices/}, 2024.
\newblock Accessed: 2024-10-01.

\bibitem[Beltagy et~al.(2020{\natexlab{a}})Beltagy, Peters, and Cohan]{Beltagy2020Longformer}
Beltagy, I., Peters, M.~E., and Cohan, A.
\newblock Longformer: The long-document transformer.
\newblock \emph{arXiv:2004.05150}, 2020{\natexlab{a}}.

\bibitem[Beltagy et~al.(2020{\natexlab{b}})Beltagy, Peters, and Cohan]{longformer}
Beltagy, I., Peters, M.~E., and Cohan, A.
\newblock Longformer: The long-document transformer.
\newblock \emph{CoRR}, abs/2004.05150, 2020{\natexlab{b}}.

\bibitem[Bisk et~al.(2020)Bisk, Zellers, Bras, Gao, and Choi]{Bisk2020}
Bisk, Y., Zellers, R., Bras, R.~L., Gao, J., and Choi, Y.
\newblock Piqa: Reasoning about physical commonsense in natural language.
\newblock In \emph{Thirty-Fourth AAAI Conference on Artificial Intelligence}, 2020.

\bibitem[Chen et~al.(2024)Chen, Sadhukhan, Ye, Zhou, Zhang, Nolte, Tian, Douze, Bottou, Jia, and Chen]{chen2024magicpiglshsamplingefficient}
Chen, Z., Sadhukhan, R., Ye, Z., Zhou, Y., Zhang, J., Nolte, N., Tian, Y., Douze, M., Bottou, L., Jia, Z., and Chen, B.
\newblock Magicpig: Lsh sampling for efficient llm generation, 2024.
\newblock URL \url{https://arxiv.org/abs/2410.16179}.

\bibitem[Child et~al.(2019)Child, Gray, Radford, and Sutskever]{child2019generatinglongsequencessparse}
Child, R., Gray, S., Radford, A., and Sutskever, I.
\newblock Generating long sequences with sparse transformers, 2019.
\newblock URL \url{https://arxiv.org/abs/1904.10509}.

\bibitem[Clark et~al.(2019)Clark, Lee, Chang, Kwiatkowski, Collins, and Toutanova]{clark2019boolq}
Clark, C., Lee, K., Chang, M.-W., Kwiatkowski, T., Collins, M., and Toutanova, K.
\newblock Boolq: Exploring the surprising difficulty of natural yes/no questions.
\newblock In \emph{Proceedings of the 2019 Conference of the North American Chapter of the Association for Computational Linguistics: Human Language Technologies, Volume 1 (Long and Short Papers)}, pp.\  2924--2936, 2019.

\bibitem[Clark et~al.(2018)Clark, Cowhey, Etzioni, Khot, Sabharwal, Schoenick, and Tafjord]{clark2018arc}
Clark, P., Cowhey, I., Etzioni, O., Khot, T., Sabharwal, A., Schoenick, C., and Tafjord, O.
\newblock Think you have solved question answering? try arc, the ai2 reasoning challenge.
\newblock \emph{arXiv preprint arXiv:1803.05457}, 2018.

\bibitem[Dao(2024)]{dao2024flashattention}
Dao, T.
\newblock Flashattention-2: Faster attention with better parallelism and work partitioning.
\newblock In \emph{The Twelfth International Conference on Learning Representations}, 2024.
\newblock URL \url{https://openreview.net/forum?id=mZn2Xyh9Ec}.

\bibitem[Dao et~al.(2022)Dao, Fu, Ermon, Rudra, and Ré]{dao2022flashattentionfastmemoryefficientexact}
Dao, T., Fu, D.~Y., Ermon, S., Rudra, A., and Ré, C.
\newblock Flashattention: Fast and memory-efficient exact attention with io-awareness, 2022.
\newblock URL \url{https://arxiv.org/abs/2205.14135}.

\bibitem[Dubey et~al.(2024)Dubey, Jauhri, Pandey, Kadian, Al-Dahle, Letman, Mathur, Schelten, Yang, Fan, et~al.]{dubey2024llama}
Dubey, A., Jauhri, A., Pandey, A., Kadian, A., Al-Dahle, A., Letman, A., Mathur, A., Schelten, A., Yang, A., Fan, A., et~al.
\newblock The llama 3 herd of models.
\newblock \emph{arXiv preprint arXiv:2407.21783}, 2024.

\bibitem[Dubois et~al.(2024)Dubois, Galambosi, Liang, and Hashimoto]{dubois2024length}
Dubois, Y., Galambosi, B., Liang, P., and Hashimoto, T.~B.
\newblock Length-controlled alpacaeval: A simple way to debias automatic evaluators.
\newblock \emph{arXiv preprint arXiv:2404.04475}, 2024.

\bibitem[Gao et~al.(2024)Gao, Tow, Abbasi, Biderman, Black, DiPofi, Foster, Golding, Hsu, Le~Noac'h, Li, McDonell, Muennighoff, Ociepa, Phang, Reynolds, Schoelkopf, Skowron, Sutawika, Tang, Thite, Wang, Wang, and Zou]{eval-harness}
Gao, L., Tow, J., Abbasi, B., Biderman, S., Black, S., DiPofi, A., Foster, C., Golding, L., Hsu, J., Le~Noac'h, A., Li, H., McDonell, K., Muennighoff, N., Ociepa, C., Phang, J., Reynolds, L., Schoelkopf, H., Skowron, A., Sutawika, L., Tang, E., Thite, A., Wang, B., Wang, K., and Zou, A.
\newblock A framework for few-shot language model evaluation, 07 2024.
\newblock URL \url{https://zenodo.org/records/12608602}.

\bibitem[{Gradient Team}(2024)]{gradient2024scaling}
{Gradient Team}.
\newblock Scaling rotational embeddings for long-context language models.
\newblock \url{https://gradient.ai/blog/scaling-rotational-embeddings-for-long-context-language-models}, May 2024.
\newblock Accessed: 2024-10-01.

\bibitem[Grattafiori et~al.(2024)]{grattafiori2024llama3herdmodels}
Grattafiori, A. et~al.
\newblock The llama 3 herd of models, 2024.
\newblock URL \url{https://arxiv.org/abs/2407.21783}.

\bibitem[Gromov et~al.(2024)Gromov, Tirumala, Shapourian, Glorioso, and Roberts]{gromov2024unreasonableineffectivenessdeeperlayers}
Gromov, A., Tirumala, K., Shapourian, H., Glorioso, P., and Roberts, D.~A.
\newblock The unreasonable ineffectiveness of the deeper layers, 2024.
\newblock URL \url{https://arxiv.org/abs/2403.17887}.

\bibitem[Gupta et~al.(2021)Gupta, Dar, Goodman, Ciprut, and Berant]{guptaMemoryefficientTransformersTopk2021}
Gupta, A., Dar, G., Goodman, S., Ciprut, D., and Berant, J.
\newblock Memory-efficient {{Transformers}} via {{Top-k Attention}}.
\newblock In \emph{Proceedings of the {{Second Workshop}} on {{Simple}} and {{Efficient Natural Language Processing}}}, pp.\  39--52, Virtual, 2021. Association for Computational Linguistics.
\newblock \doi{10.18653/v1/2021.sustainlp-1.5}.

\bibitem[Hendrycks et~al.(2020)Hendrycks, Burns, Basart, Zou, Mazeika, Song, and Steinhardt]{hendrycks2020measuring-MMLU}
Hendrycks, D., Burns, C., Basart, S., Zou, A., Mazeika, M., Song, D., and Steinhardt, J.
\newblock Measuring massive multitask language understanding.
\newblock In \emph{International Conference on Learning Representations}, 2020.

\bibitem[Hsieh et~al.(2024)Hsieh, Sun, Kriman, Acharya, Rekesh, Jia, Zhang, and Ginsburg]{hsiehRULERWhatsReal2024}
Hsieh, C.-P., Sun, S., Kriman, S., Acharya, S., Rekesh, D., Jia, F., Zhang, Y., and Ginsburg, B.
\newblock {{RULER}}: {{What}}'s the {{Real Context Size}} of {{Your Long-Context Language Models}}?, April 2024.

\bibitem[Kamradt(2023)]{kamradt2023needle}
Kamradt, G.
\newblock Needle in a haystack - pressure testing llms.
\newblock \url{https://github.com/gkamradt/LLMTest}, 2023.
\newblock GitHub repository.

\bibitem[Keisuke et~al.(2019)Keisuke, Ronan, Chandra, and Yejin]{ai2:winogrande}
Keisuke, S., Ronan, L.~B., Chandra, B., and Yejin, C.
\newblock Winogrande: An adversarial winograd schema challenge at scale.
\newblock 2019.

\bibitem[Kwon et~al.(2023)Kwon, Li, Zhuang, Sheng, Zheng, Yu, Gonzalez, Zhang, and Stoica]{kwon2023efficient}
Kwon, W., Li, Z., Zhuang, S., Sheng, Y., Zheng, L., Yu, C.~H., Gonzalez, J.~E., Zhang, H., and Stoica, I.
\newblock Efficient memory management for large language model serving with pagedattention.
\newblock In \emph{Proceedings of the ACM SIGOPS 29th Symposium on Operating Systems Principles}, 2023.

\bibitem[Liu et~al.(2024{\natexlab{a}})Liu, Chen, Lu, Jiang, Han, Zhang, Chen, Zhang, Ding, Zhang, Chen, Yang, Yang, and Qiu]{liu2024retrievalattentionacceleratinglongcontextllm}
Liu, D., Chen, M., Lu, B., Jiang, H., Han, Z., Zhang, Q., Chen, Q., Zhang, C., Ding, B., Zhang, K., Chen, C., Yang, F., Yang, Y., and Qiu, L.
\newblock Retrievalattention: Accelerating long-context llm inference via vector retrieval, 2024{\natexlab{a}}.
\newblock URL \url{https://arxiv.org/abs/2409.10516}.

\bibitem[Liu et~al.(2023)Liu, Zaharia, and Abbeel]{liu2023ringattentionblockwisetransformers}
Liu, H., Zaharia, M., and Abbeel, P.
\newblock Ring attention with blockwise transformers for near-infinite context, 2023.
\newblock URL \url{https://arxiv.org/abs/2310.01889}.

\bibitem[Liu et~al.(2024{\natexlab{b}})Liu, Yan, Zaharia, and Abbeel]{liu2024worldmodelmillionlengthvideo}
Liu, H., Yan, W., Zaharia, M., and Abbeel, P.
\newblock World model on million-length video and language with blockwise ringattention, 2024{\natexlab{b}}.
\newblock URL \url{https://arxiv.org/abs/2402.08268}.

\bibitem[Malkov \& Yashunin(2020)Malkov and Yashunin]{10.1109/TPAMI.2018.2889473}
Malkov, Y.~A. and Yashunin, D.~A.
\newblock Efficient and robust approximate nearest neighbor search using hierarchical navigable small world graphs.
\newblock \emph{IEEE Trans. Pattern Anal. Mach. Intell.}, 42\penalty0 (4):\penalty0 824–836, April 2020.
\newblock ISSN 0162-8828.
\newblock \doi{10.1109/TPAMI.2018.2889473}.
\newblock URL \url{https://doi.org/10.1109/TPAMI.2018.2889473}.

\bibitem[Mihaylov et~al.(2018)Mihaylov, Clark, Khot, and Sabharwal]{OpenBookQA2018}
Mihaylov, T., Clark, P., Khot, T., and Sabharwal, A.
\newblock Can a suit of armor conduct electricity? a new dataset for open book question answering.
\newblock In \emph{EMNLP}, 2018.

\bibitem[Pope et~al.(2023)Pope, Douglas, Chowdhery, Devlin, Bradbury, Heek, Xiao, Agrawal, and Dean]{pope2023efficiently}
Pope, R., Douglas, S., Chowdhery, A., Devlin, J., Bradbury, J., Heek, J., Xiao, K., Agrawal, S., and Dean, J.
\newblock Efficiently scaling transformer inference.
\newblock \emph{Proceedings of Machine Learning and Systems}, 5:\penalty0 606--624, 2023.

\bibitem[Rajpurkar et~al.(2016)Rajpurkar, Zhang, Lopyrev, and Liang]{rajpurkar2016squad100000questionsmachine}
Rajpurkar, P., Zhang, J., Lopyrev, K., and Liang, P.
\newblock Squad: 100,000+ questions for machine comprehension of text, 2016.
\newblock URL \url{https://arxiv.org/abs/1606.05250}.

\bibitem[Sheng et~al.(2023)Sheng, Zheng, Yuan, Li, Ryabinin, Chen, Liang, R\'{e}, Stoica, and Zhang]{sheng2023flexgen}
Sheng, Y., Zheng, L., Yuan, B., Li, Z., Ryabinin, M., Chen, B., Liang, P., R\'{e}, C., Stoica, I., and Zhang, C.
\newblock Flexgen: high-throughput generative inference of large language models with a single gpu.
\newblock In \emph{Proceedings of the 40th International Conference on Machine Learning}, ICML'23. JMLR.org, 2023.

\bibitem[Singhania et~al.(2024)Singhania, Singh, He, Feizi, and Bhatele]{singhania2024lokilowrankkeysefficient}
Singhania, P., Singh, S., He, S., Feizi, S., and Bhatele, A.
\newblock Loki: Low-rank keys for efficient sparse attention, 2024.
\newblock URL \url{https://arxiv.org/abs/2406.02542}.

\bibitem[Tang et~al.(2024)Tang, Zhao, Zhu, Xiao, Kasikci, and Han]{tang2024questqueryawaresparsityefficient}
Tang, J., Zhao, Y., Zhu, K., Xiao, G., Kasikci, B., and Han, S.
\newblock Quest: Query-aware sparsity for efficient long-context llm inference, 2024.
\newblock URL \url{https://arxiv.org/abs/2406.10774}.

\bibitem[Touvron et~al.(2023{\natexlab{a}})Touvron, Lavril, Izacard, Martinet, Lachaux, Lacroix, Rozi{\`e}re, Goyal, Hambro, Azhar, et~al.]{touvron2023llama}
Touvron, H., Lavril, T., Izacard, G., Martinet, X., Lachaux, M.-A., Lacroix, T., Rozi{\`e}re, B., Goyal, N., Hambro, E., Azhar, F., et~al.
\newblock Llama: Open and efficient foundation language models.
\newblock \emph{arXiv preprint arXiv:2302.13971}, 2023{\natexlab{a}}.

\bibitem[Touvron et~al.(2023{\natexlab{b}})Touvron, Martin, Stone, Albert, Almahairi, Babaei, Bashlykov, Batra, Bhargava, Bhosale, et~al.]{touvron2023llama2}
Touvron, H., Martin, L., Stone, K., Albert, P., Almahairi, A., Babaei, Y., Bashlykov, N., Batra, S., Bhargava, P., Bhosale, S., et~al.
\newblock Llama 2: Open foundation and fine-tuned chat models.
\newblock \emph{arXiv preprint arXiv:2307.09288}, 2023{\natexlab{b}}.

\bibitem[Xiao et~al.(2023)Xiao, Tian, Chen, Han, and Lewis]{xiao2023streamingllm}
Xiao, G., Tian, Y., Chen, B., Han, S., and Lewis, M.
\newblock Efficient streaming language models with attention sinks.
\newblock \emph{arXiv}, 2023.

\bibitem[Xiao et~al.(2024)Xiao, Tian, Chen, Han, and Lewis]{streamingllm}
Xiao, G., Tian, Y., Chen, B., Han, S., and Lewis, M.
\newblock Efficient streaming language models with attention sinks.
\newblock In \emph{{ICLR}}. OpenReview.net, 2024.

\bibitem[Yang et~al.(2018)Yang, Qi, Zhang, Bengio, Cohen, Salakhutdinov, and Manning]{yang2018hotpotqadatasetdiverseexplainable}
Yang, Z., Qi, P., Zhang, S., Bengio, Y., Cohen, W.~W., Salakhutdinov, R., and Manning, C.~D.
\newblock Hotpotqa: A dataset for diverse, explainable multi-hop question answering, 2018.
\newblock URL \url{https://arxiv.org/abs/1809.09600}.

\bibitem[Zellers et~al.(2019)Zellers, Holtzman, Bisk, Farhadi, and Choi]{zellers-etal-2019-hellaswag}
Zellers, R., Holtzman, A., Bisk, Y., Farhadi, A., and Choi, Y.
\newblock {H}ella{S}wag: Can a machine really finish your sentence?
\newblock In \emph{Proceedings of the 57th Annual Meeting of the Association for Computational Linguistics}, pp.\  4791--4800, Florence, Italy, July 2019. Association for Computational Linguistics.
\newblock \doi{10.18653/v1/P19-1472}.
\newblock URL \url{https://aclanthology.org/P19-1472}.

\bibitem[Zhang et~al.(2024)Zhang, Ji, Chen, Fu, Miao, Nie, Chen, and Cui]{zhang2024pqcacheproductquantizationbasedkvcache}
Zhang, H., Ji, X., Chen, Y., Fu, F., Miao, X., Nie, X., Chen, W., and Cui, B.
\newblock Pqcache: Product quantization-based kvcache for long context llm inference, 2024.
\newblock URL \url{https://arxiv.org/abs/2407.12820}.

\bibitem[Zhang et~al.(2023)Zhang, Sheng, Zhou, Chen, Zheng, Cai, Song, Tian, Ré, Barrett, Wang, and Chen]{zhang2023h2oheavyhitteroracleefficient}
Zhang, Z., Sheng, Y., Zhou, T., Chen, T., Zheng, L., Cai, R., Song, Z., Tian, Y., Ré, C., Barrett, C., Wang, Z., and Chen, B.
\newblock H$_2$o: Heavy-hitter oracle for efficient generative inference of large language models, 2023.
\newblock URL \url{https://arxiv.org/abs/2306.14048}.

\bibitem[Zheng et~al.(2023)Zheng, Chiang, Sheng, Zhuang, Wu, Zhuang, Lin, Li, Li, Xing, et~al.]{zheng2023judging}
Zheng, L., Chiang, W.-L., Sheng, Y., Zhuang, S., Wu, Z., Zhuang, Y., Lin, Z., Li, Z., Li, D., Xing, E., et~al.
\newblock Judging llm-as-a-judge with mt-bench and chatbot arena.
\newblock \emph{Advances in Neural Information Processing Systems}, 36:\penalty0 46595--46623, 2023.

\end{thebibliography}
